\documentclass[10pt,journal,compsoc]{IEEEtran}
\ifCLASSOPTIONcompsoc
\usepackage[nocompress]{cite}
\else
\usepackage{cite}
\fi

\ifCLASSINFOpdf
\usepackage[pdftex]{graphicx}
\usepackage{multirow}
\usepackage{amsmath}
\usepackage{amssymb}
\usepackage{caption}

\usepackage{color} 
\usepackage{makecell}
\usepackage{pifont}
\usepackage[colorlinks,linkcolor=red]{hyperref}
\usepackage{verbatim} 
\usepackage{comment}
\usepackage{booktabs}
\usepackage{subfigure}
\usepackage{ragged2e}
\usepackage{stfloats}   % 控制跨栏图像出现在当前页面顶部
\usepackage[table,xcdraw]{xcolor}   % 表格上色

\else
\fi

\begin{document}
%
% paper title
% Titles are generally capitalized except for words such as a, an, and, as,
% at, but, by, for, in, nor, of, on, or, the, to and up, which are usually
% not capitalized unless they are the first or last word of the title.
% Linebreaks \\ can be used within to get better formatting as desired.
% Do not put math or special symbols in the title.
\title{PanMatch: Unleashing the Potential of Large Vision Models for Unified Matching Models}

\author{Yongjian Zhang, Longguang Wang, Kunhong Li, Ye Zhang, Yun Wang, Liang Lin, \emph{Fellow, IEEE}, Yulan Guo*% <-this % stops a space
    \IEEEcompsocitemizethanks{\IEEEcompsocthanksitem 
    \textcolor{black}{
        *Corresponding author: Yulan Guo.
        \IEEEcompsocthanksitem Yongjian Zhang, Longguang Wang, Kunhong Li, Ye Zhang and Yulan Guo are with the School of Electronics and Communication Engineering, the Shenzhen Campus of Sun Yat-sen University, Sun Yat-sen University, Shenzhen, China (e-mail: 
            zhangyj85@mail2.sysu.edu.cn,
            wanglongguang15@nudt.edu.cn, 
            likh25@mail2.sysu.edu.cn,
            zhangy2658@ sysu.edu.cn,
            guoyulan@sysu.edu.cn). %\protect\\
        \IEEEcompsocthanksitem Liang Lin is with the School of Computer Science and  Engineering, Sun Yat-sen University, Guangzhou, China  (e-mail: linliang@ieee.org).
        Yun Wang is with the Department of Computer Science, City University of Hong Kong, Kowloon 999077, Hong Kong SAR, China (e-mail: ywang3875-c@my.cityu.edu.hk).
    }
    }
    % \thanks{Manuscript received XX XX, 2020; revised XX XX, 2020.}
}

% The paper headers
% \markboth{IEEE Transactions on Pattern Analysis and Machine Intelligence,~Vol.~XX, No.~XX, XX~2025}%
\markboth{JOURNAL OF \LaTeX ~CLASS FILES,~Vol.~XX, No.~XX, July~2025}%
{Zhang \MakeLowercase{\textit{et al.}}: PanMatch: Unleashing the Potential of Large Vision Models for Zero-shot Matching Tasks}

% for Computer Society papers, we must declare the abstract and index terms
% PRIOR to the title within the \IEEEtitleabstractindextext IEEEtran
% command as these need to go into the title area created by \maketitle.
% As a general rule, do not put math, special symbols or citations
% in the abstract or keywords.
\IEEEtitleabstractindextext{%
    \begin{abstract}
        \justifying   			
        This work presents \emph{PanMatch}, a versatile foundation model for robust correspondence matching. 
        % 和其他工作相比 (RVC model, unimatch, RGM), 独特之处
            Unlike previous methods that rely on task-specific architectures and domain-specific fine-tuning to support tasks like stereo matching, optical flow or feature matching,
            our key insight is that any two-frame correspondence matching task can be addressed within a 2D displacement estimation framework using the same model weights.
            Such a formulation eliminates the need for designing specialized unified architectures or task-specific ensemble models. 
            Instead, it achieves multi-task integration {by endowing displacement estimation algorithms with unprecedented generalization capabilities}.
        % 我们的具体做法
            To this end, we highlight the importance of a robust feature extractor applicable across multiple domains and tasks, 
            and propose the \emph{feature transformation pipeline} that leverage all-purpose features from Large Vision Models to endow matching baselines with zero-shot cross-view matching capabilities.
            Furthermore, we assemble a cross-domain dataset with near \emph{1.8 million samples} from stereo matching, optical flow, and feature matching domains to pretrain PanMatch.
        % 我们的优势
            We demonstrate the versatility of PanMatch across a wide range of domains and downstream tasks using \emph{the same model weights}.
        % 我们取得的效果
            Our model outperforms UniMatch and Flow-Anything on cross-task evaluations, and achieves comparable performance to most state-of-the-art task-specific algorithms on task-oriented benchmarks.
            Additionally, PanMatch presents unprecedented zero-shot performance in abnormal scenarios, such as rainy day and satellite imagery, where most existing robust algorithms fail to yield meaningful results.
    \end{abstract}
    
    % Note that keywords are not normally used for peerreview papers.
    \begin{IEEEkeywords}
        Stereo Matching, Optical Flow Estimation, Feature Matching, Depth Estimation, Domain Generalization.
\end{IEEEkeywords}}

% make the title area
\maketitle

% To allow for easy dual compilation without having to reenter the
% abstract/keywords data, the \IEEEtitleabstractindextext text will
% not be used in maketitle, but will appear (i.e., to be "transported")
% here as \IEEEdisplaynontitleabstractindextext when the compsoc 
% or transmag modes are not selected <OR> if conference mode is selected 
% - because all conference papers position the abstract like regular
% papers do.
\IEEEdisplaynontitleabstractindextext
% \IEEEdisplaynontitleabstractindextext has no effect when using
% compsoc or transmag under a non-conference mode.

% For peer review papers, you can put extra information on the cover
% page as needed:
% \ifCLASSOPTIONpeerreview
% \begin{center} \bfseries EDICS Category: 3-BBND \end{center}
% \fi
%
% For peerreview papers, this IEEEtran command inserts a page break and
% creates the second title. It will be ignored for other modes.
\IEEEpeerreviewmaketitle

\IEEEraisesectionheading{\section{Introduction}}

\IEEEPARstart{F}{inding} correspondences in viewpoint-overlapping images is one of the key ways to achieve 3D scene perception and reconstruction. This technique serves as the foundation for various real-world applications, including stereo matching for driving and navigation, optical flow for video editing and action recognition, and feature matching for 3D reconstruction.
	
% 三个任务的区别与联系
Previous research developed specialized architectures and model weights for specific correspondence tasks due to significant difference in task settings, as outlined in Table~\ref{tab:task_comparison}.
For instance, stereo matching operates on a pair of synchronized, rectified images and identifies correspondences along horizontal epipolar lines.
Feature matching focus on finding reliable correspondence of rigid scenes from varying camera poses and times.
Optical flow estimates pixel-wise {displacements} in dynamic scenes over consecutive frames.
% 单独处理任务的好处和坏处
Using task-specific priors to construct models simplifies network design and enhances inference efficiency.
However, these individual pipelines inherently limits the adaptability of the algorithms across tasks, resulting in numerous specialized architectures and weights for different scenarios, which complicates real-world deployment.
Additionally, robust geometric priors learned from diverse data sources cannot be seamlessly integrated into a single model during training.

Given the inherent similarities across stereo matching, optical flow, and feature matching, several efforts have been made to address correspondence matching within a unified architecture~\cite{unimatch,RGM}. 
% Xu \textit{et al.} \cite{unimatch} leverage Transformer for discriminative feature representations and find correspondeces by comparing feature similarities, which formulate a unified architecture for dense correspondence matching. 
% Zhang \textit{et al.} \cite{RGM} leverage a optical flow estimation network to first obtain the dense matching resutls and then filter out the unreliable pairs, which is also a unified architecture for multiple matching tasks.
However, due to task- and domain-specific biases, directly applying these methods across different tasks results in relatively low performance. This raises a question: Can we develop a truly unified model for all correspondence tasks using the same weights?

% 说明是什么原因导致目前无法使用同一的模型解决多任务多场景?
Drawing from the success of large vision models in other fields~\cite{vggt}, we summarize the major challenges of developing such a unified model as twofold.
(i) Although most correspondence matching models follow a similar four-step pipeline—feature extraction, cross-view correlation, global cost optimization, and target regression—most methods encode task-specific geometric inductive biases to formulate task-dependent cost volumes~\cite{unimatch}, followed by target-dependent aggregation networks for regression. This task-relevant design and inconsistent output targets hinder the unification of different correspondence-matching tasks into a single architecture.
(ii) Previous methods are pretrained on task-specific and scale-limited datasets, which inevitably leads to domain-sensitive representations and limited generalization capability~\cite{ITSA}.
% This, in turn, impairs both their sim-to-real generalization performance and their ability to adapt across tasks.
{Although several works augment their generalization through mixed training datasets~\cite{RGM}, they still rely on task-oriented samples, failing to meet the in-the-wild demands of cross-domain and cross-task applications.}

% In this paper, we aim at developing a single unified model and using the same checkpoint to solve any correspondence tasks, including stereo matching, optical flow and feature matching, as shown in Fig.~\ref{}. 
% The main challenges for achieving a truth unified model lies in two aspects: a robust feature extractor and a general global matcher.
% On one hand, previous methods pretrain on the task-specific scale-limited datasets, which inevitably encodes the shortcuts and task-sensitive representations into the feature extractor, therefore hinder their sim-to-real generalization performance~\cite{ITSA, HVT}, as well as their cross-task adaptation~\cite{unimatch}.
% Although some works expend their training samples with mixture datasets~\cite{CFNet, iRAFT, CREStereo++, LoS}, their available training samples are still limited, and face the generalization disaster to some unseen scenes.
% On the other hand, most methods encode the task-specific geometric inductive bias~\cite{unimatch} to formulate the task-dependent cost volumes, and leverage vary aggregation networks for cost filtering. Such a pipeline also introduces additional challenges in unifying all correspondence matching tasks into a single architecture.

% 我们的解决方案是是什么, 受到什么启发
To address the first challenge, we construct a unified model by removing task-specific geometric priors and relaxing the problem to a pure all-pairs correspondence matching task, {which can be implemented using standard optical flow estimation approaches.} 
% (\textit{i.e.}, flow estimation). 
By this way, the core challenge of developing a unified architecture shifts toward {endowing an optical flow baseline with generalization capabilities across tasks and domains}.
% Additionally, we introduce a sub-network to assess estimation confidence, which filters out unreliable results for feature matching. 
% The overall framework centers on optical flow estimation, for which we select an effective optical flow algorithm as the solution.
% To this end, %starting with an existing robust optical flow algorithm, 
{To this end, we emphasize the importance of \textit{robust feature representations}, and leverage large vision models (LVMs) as a feasible solution.
We design a feature transformation pipeline, which transfer the multi-layer all-purpose features extracted by LVM into matching-specific representations.
This is achieved through a guided feature upsampling block for detail restoration and a lightweight fusion adapter for feature transfer.
By integrating these designs, the optical flow baseline can be extended to a unified model capable of both cross-domain generalization and cross-task adaptability.} % and {\textit{scalable training paradigm}}.
\begin{table}[t]
    \setlength{\tabcolsep}{3pt}
    \centering
    \caption{Comparisons on different correspondence matching tasks.}
    \begin{tabular}{lccc}
    \hline
    Task             & Focus                    & Image Pair            & Target                \\ \hline
    Stereo Matching  & Static Scenes            & Similar               & \textbf{Dense}        \\
    Feature Matching & Rigid Scenes           & \textbf{Varying}      & Semi-Dense            \\
    Optical Flow     & \textbf{Dynamic Scenes}  & Similar               & \textbf{Dense}        \\ \hline
    \end{tabular}
    \label{tab:task_comparison}
\end{table}

% 我们的统一框架 overview
\begin{figure}[t]
    \centering
    \includegraphics[width=0.45\textwidth]{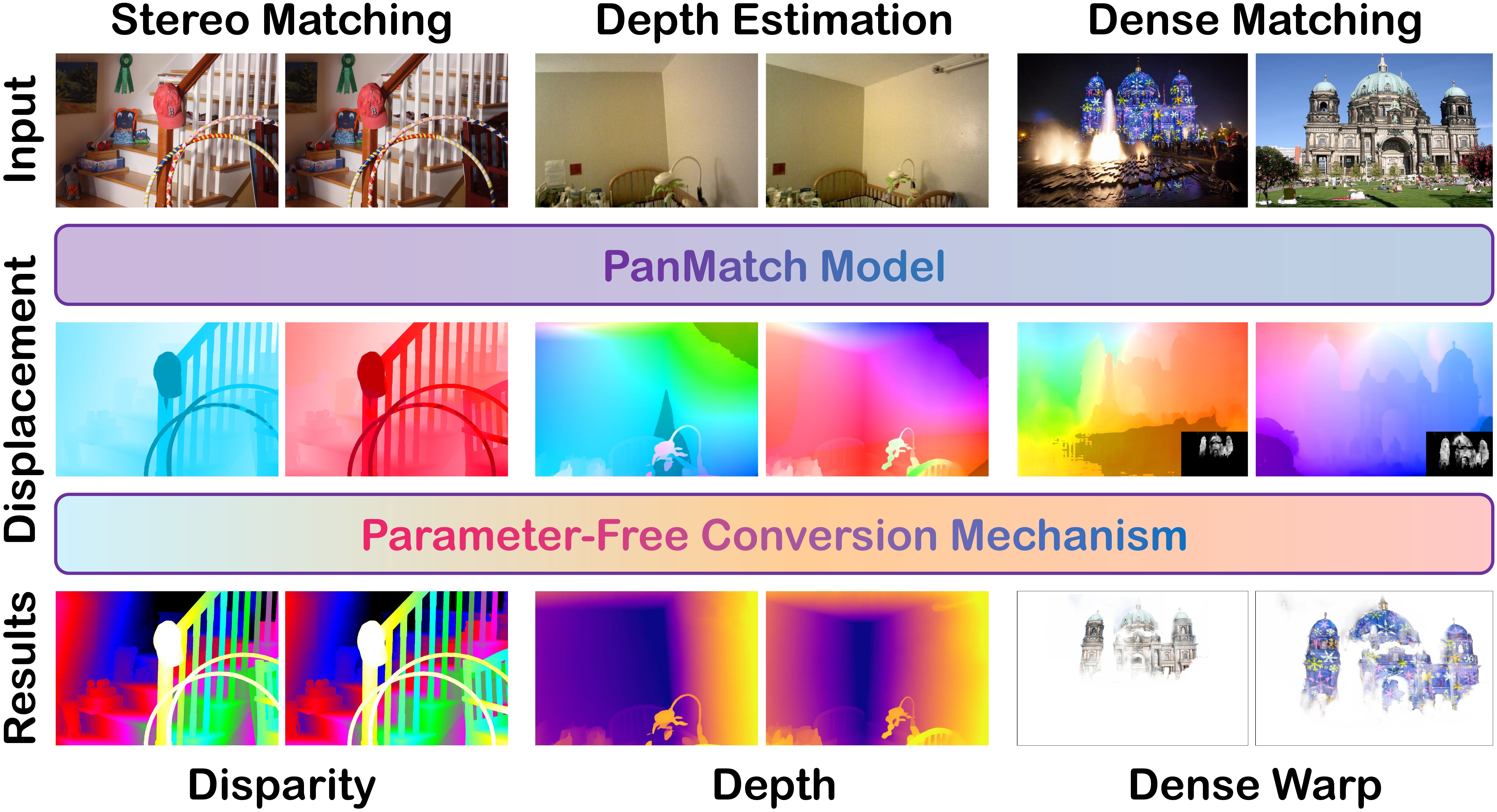}
    \caption{An overview of the proposed unified architecture. We first estimate the reliable 2D displacement field with PanMatch, and then derive the target correspondence format by parameter-free calculations.}
    \label{fig:overview}
\end{figure}

% 我们如何实现多个任务的统一
% 仅使用光流的范式和光流数据是远远不够的, 从数据多样性(比如无法适配stereo的booster)和任务有效性(比如无法适配megadepth)的角度探讨
% Previous researches only focus on the design of unified architectures while omitting the contributions of diverse training samples, leading sub-optimal performance on the real-world downstream applications with unseen flow distributions. 
% Our unified model also offers another advantage: it can effectively utilize data from multiple tasks and sources for training, a feature that other unified models cannot support.
% Our another key insight is that any correspondence matching task can be unified as the optical flow formulation, once the feature extractor is robust to domain changes. 

{To address the second challenge, we {reorganize} various correspondence matching datasets to support our unified matching formulation, thereby expanding the scale and diversity of available training data.}
% With our proposed Feature Transformation module, the selected optical flow baseline is capable to be applied across domains to estimate accurate optical flow, which can be further transformed into disparity, depth, or warping correspondences through parameter-free calculations, as illustrated in Fig.~\ref{fig:overview} and demonstrated in Sec.~\ref{sec:unify_model}.
% Furthermore, since the flow-to-targets calculations are reversible, we can convert various correspondence matching datasets into optical flow formulations, thereby expanding the scale and diversity of available training data.
% With our unified formulation in Fig.~\ref{fig:overview}, we demonstrate that results such as disparity, depth, and warping correspondences can be derived from the estimated optical flow using parameter-free and reversible calculations.
% Furthermore, benefit from the reversable calculations, we can transform various correspondence matching datasets into optical flow formulations, thereby expanding the scale and diversity of available training data.
In practice, we collect nearly 1.8 million samples from existing datasets developed for stereo matching~\cite{crestereo, fallingthings}, optical flow~\cite{DispNetC, FlowNet, autoflow, vkitti2, tartanair} and depth estimation~\cite{hypersim, megadepth}. 
Note that, our dataset is of higher quality and diversity compared with croco v2~\cite{croco_v2}, StereoAnything~\cite{StereoAnything} and FoundationStereo~\cite{FoundationStereo}. 
%our dataset provides higher-resolution image pairs with annotations. 
%the quality of our samples is much croco v2~\cite{croco_v2}, which only consists of low-resolution image pairs without annotations.}
% The scale of our pretraining samples is quite different to unimatch~\cite{unimatch}, which relies on a task-specific feature alignment method to encode geometric inductive biases, thereby limiting its ability to learn from cross-task data.
%Meanwhile, the data diversity also differs from StereoAnything~\cite{StereoAnything} and FoundationStereo~\cite{FoundationStereo}, as both of which only collect and generate stereo datasets for training.
% Notably, our formulation is general and updatable, allowing for the inclusion of new optical flow algorithms and correspondence datasets for training.
{Moreover, as our data {reorganization} strategy is applicable to crowd-sourced matching datasets, it can incorporate newer datasets mentioned above to achieve even larger scales for learning correspondence matching.}

% 我们取得的总体效果
We have experimentally demonstrated that leveraging representations from LVMs enhances the cross-domain generalization ability of unified models. 
Furthermore, the integration of large-scale multi-task training data substantially improves model adaptability across diverse tasks.
Based on our unified task formulation and dual advances in model design and data strategy, we develop PanMatch, a versatile foundation model that achieves state-of-the-art performance across multiple benchmarks using identical model weights, as illustrated in Fig.~\ref{fig:overview}.
{Our PanMatch achieves the best performance on ETH3D and Spring benchmarks, and ranks second on Middlebury, Sintel and WxBS benchmark when compared to task-specific robust algorithms, demonstrating strong cross-task generalization.}
% this module and a selected optical flow baseline, we introduce a unified model, PanMatch, designed for any correspondence matching task.
% Our PanMatch outperforms UniMatch~\cite{unimatch} and RoMa~\cite{RoMa}, while achieving the state-of-the-art performance across multiple task benchmarks.

% 匹配的应用

% 匹配存在的域泛化问题
% 立体匹配中域泛化研究取得的成功
% 其他光流任务中域泛化方法的主流思想
% 使用大模型的动机
% 使用大模型存在的问题
% 我们的对应策略

% 我们的主要贡献
This work significantly extends our previous ECCV 2024 conference paper, \textit{FormerStereo}~\cite{FormerStereo}, with the following new contributions:
% \begin{itemize}
%     % insights 上的不同
%     \item The original \textit{FormerStereo}~\cite{FormerStereo} focuses on enhancing the generalization capacity of existing stereo matching baselines using LVMs, whereas this paper proposes a more general framework that unlocks the potential of LVMs for a variety of correspondence tasks.
%     % tasks 上的不同
%     \item We highlight the challenges of utilizing LVMs for correspondence matching and improve upon the \textit{FormerStereo} framework. Additionally, we extend its applicability to optical flow tasks.
%     % 效果上/结果上的不同
%     \item We propose \textit{PanMatch}, a versatile foundation model capable of addressing diverse correspondence matching tasks, including stereo matching, optical flow estimation, feature matching and depth estimation.
%     \item Our PanMatch achieves comparable or even the best performance across multiple tasks, and the baselines integrated with our Feature Transformation module achieve new state-of-the-art generalization performance.
% \end{itemize}
\begin{itemize}
    \item {We analyze the challenges of unifying diverse correspondence matching tasks and summarize the limitations of existing unified methods.}
    % 
    % \item {We propose to unify any two-view matching task with all-pairs pixel matching paradigm, and implement it by inheriting optical flow baselines.}
    \item {We propose to unify any two-view matching task as all-pairs pixel matching paradigm, and extend FormerStereo to optical flow baselines to develop a unified matching model.}
    % framework 作用范围的不同
    \item {We emphasize the importance of large-scale, high-quality training data, and collect 1.8 million cross-task samples for pretraining.}
    % \item {We extend FormerStereo to develop a unified model for multiple tasks.
    % Besides, we collect 1.8 million cross-task samples for pretraining. Both strategies facilitate the construction and generalization of a unified model.}
    \item {We introduce \emph{PanMatch}, \emph{the first} versatile foundation model capable of addressing diverse correspondence matching tasks, including stereo matching, optical flow estimation, feature matching and depth estimation. Experimental results demonstrate its cross-domain and cross-task zero-shot capabilities, achieving performance comparable to task-specific state-of-the-art (SOTA) methods.}
\end{itemize}

%---------------------------------------------------------------------------
\section{Related Work}
% 插入已有工作的共性问题, 我们从各个独立任务展开相关工作概述, 并从xx角度探讨存在的问题

% In this section, we XXXX.
In this section, we focus on three matching tasks, including stereo matching (Sec.~\ref{sec:stereo_matching}), optical flow estimation (Sec.~\ref{sec:optical_flow}) and feature matching (Sec.~\ref{sec:feature_matching}). We also review the applications of LVM in two-view tasks (Sec.~\ref{sec:LVM_application}).

\subsection{Stereo Matching}
\label{sec:stereo_matching}
% filter based methods
In the past decade, learning-based methods have gradually become the dominant approach in stereo matching~\cite{stereo_survey, stereo_survey2}.
Early techniques~\cite{MC-CNN, SGM-Net} replace partial steps in the traditional stereo matching pipeline~\cite{stereo-taxonomy} with networks to alleviate the reliance on handcrafted optimization~\cite{SGM}. Since the success of DispNet~\cite{DispNetC}, end-to-end networks with data-driven optimization have gained popularity, with subsequent research focusing on enhancing performance and efficiency through advanced model designs. These approaches primarily improve cost aggregation algorithms~\cite{PSMNet, GANet, AANet, ACVNet} or disparity refinement networks~\cite{iresnet++, HSMNet, Fast-ACVNet, HITNet, IGEV++} for in-domain performance, while overlooking cross-domain applicability.
% with better network architectures to increase accuracy , or reducing redundancy and leveraging lightweight designs to achieve fast inference \cite{AANet, HSMNet, Fast-ACVNet, HITNet}.
% However, the in-domain performance of these methods is highly dependent on the training data and influenced by domain bias, restricting their application without target-domain fine-tuning.
% iterative based methods
Motivated by the impressive generalization capabilities of RAFT~\cite{RAFT} in optical flow estimation, Lipson \textit{et al.}~\cite{RAFT-Stereo} adapted this technique for stereo matching. 
Building on the RAFT architecture, subsequent approaches~\cite{crestereo, IGEV, Selective-Stereo} mainly focus on enhancing context understanding and accelerating iterative convergence. Nevertheless, the generalization performance of these methods heavily depends on the seen context, which may fail in unseen domains.
% To name a list, Xu \textit{et al.} \cite{IGEV} combine the geometry priors in group volume into the hiddent state to improve the textless regions. 
% Wang \textit{et al.} \cite{Selective-Stereo} introduce the contextual spatial attention to enhance the context representations, following a modified GRU module to enlarge the receptive field of textureless regions, so that enhance the xxx performance.
% However, their generalization performance highly rely on the seen context, which maybe fail in the unseen domain even in scenes with rich geometric clues, as shown in Fig.~\ref{}. 

% generalization methods
Alongside efforts to fine-tune models for target domains, some research has aimed to improve the generalization performance to unseen domains without fine-tuning.
Zhang \textit{et al.}~\cite{DSMNet} suggested that learning domain-invariant representations is crucial for generalization. To this end, they proposed domain normalization and SGF layers to reduce domain-sensitive local details.
Cai \textit{et al.}~\cite{MS-Net} employed handcrafted descriptors instead of trainable feature extractors to avoid domain-sensitive information. 
Liu \textit{et al.} \cite{GraftNet} {employed frozen, broad-spectrum, task-oriented features to construct a general cosine similarity cost space, thereby mitigating the influence of feature variability on the matching distribution.}
Zhang \textit{et al.} \cite{FCStereo} utilized contrastive learning to enhance feature similarity between matching pixels while penalizing mismatched regions to avoid ambiguous feature representations. 
Chuah \textit{et al.} \cite{ITSA, ITSA_PAMI} argued that shortcut information hinders generalization. Therefore, they obtain shortcut-invariant features by enhancing representation consistency learning for shortcut perturbation. 
Chang \textit{et al.} \cite{HVT} further proposed a hierarchical visual transformation training framework with adversarial learning to mitigate shortcuts. 
Rao \textit{et al.} \cite{MaskNet} adopted mask representation learning as an auxiliary task for the feature extractor in stereo matching, improving generalization performance and stability.
% Meanwhile, maintaining cross-view feature consistency between matching pixels in different domains is essential for generalizaed stereo matching. 
These methods have advanced the development of generalized stereo matching algorithms by emphasizing robust feature representations.
However, learning domain-invariant features from limited data remains a significant challenge, and the additional modules and multitask settings inevitably impacts the in-domain accuracy of baseline methods.

\subsection{Optical Flow}
\label{sec:optical_flow}
% 主要涉及三个工作, FlowNet, RAFT, FlowFormer & GMFlow
Optical flow refers to the 2D motion of pixels from one frame to the next in a video. 
Traditional methods~\cite{ProbFlow} regard this task as an energy minimization problem, utilizing preset human-defined priors to obtain general solutions.
With the emergence of deep learning, model design and data collection have replaced handcrafted priors and features, becoming central to performance improvements.
Over the past decade, 
Dosovitskiy \textit{et al.} \cite{FlowNet} pioneered an end-to-end CNN network to directly regress optical flow within a data-driven pipeline, which is followed by subsequent works \cite{DispNetC,PWC-Net,FlowNet2.0}.
Recently, Teed \textit{et al.} \cite{RAFT} introduced RAFT, a novel architecture that iteratively updates flow estimations to achieve generalizable estimation, setting a new baseline to the community \cite{GMA, SEA-RAFT}.
Nonetheless, these methods still struggle with small and fast-moving objects, which require global correlation rather than local cost volumes. 
To overcome this limitation, Xu \textit{et al.} \cite{GMFlow, unimatch} leveraged Transformers to develop a robust feature extractor, enabling the direct estimation of optical flow from an all-pairs correlation volume.
Huang \textit{et al.} \cite{FlowFormer} processed the all-pairs correlation volume with Transformers to obtain a global cost memory, providing global cues for flow estimations.
Since RAFT, generalization performance has been increasingly valued, yet most methods still heavily rely on fine-tuning to achieve optimal performance on the Sintel~\cite{Sintel} and KITTI~\cite{KITTI2015} benchmarks.

% 再讲一讲光流的数据集, 各个数据集的提出来的目的
The data-driven learning paradigm has also spawned numerous open-source optical flow datasets.
Dosovitskiy \textit{et al.} \cite{FlowNet} proposed the first large-scale FlyingChair dataset. Then, Mayer \textit{et al.} \cite{DispNetC} introduced a more realistic FlyingThings dataset. 
% Sun \textit{et al.} \cite{autoflow} constructed an optimization pipeline to synthetic pretraining examples for Sintel~\cite{Sintel}.
{Sun \textit{et al.} \cite{autoflow} proposed a self-optimizing data generation pipeline that synthesizes pretraining data to boost baseline performance on the Sintel~\cite{Sintel} benchmark.}
Additionally, researchers in the fields like SLAM~\cite{tartanair}, autonomous driving~\cite{vkitti2, viper,KITTI2012,KITTI2015}, and data generation~\cite{kubric} create datasets with additional optical flow annotation, significantly increasing the quantity of training data for optical flow estimation.
However, existing optical flow annotations focus on continuous motion of dynamic objects, ignoring the complex real-world scenarios, such as non-Lambertian reflections and significant differences in viewing angles caused by high-speed camera motion.

\subsection{Feature Matching}
\label{sec:feature_matching}
% 分为 detection-rely 和 detector-free 两类, 其中 detector-free 的方法重点讲那些能够输出 dense output 的方法, 包含带置信度估计的光流方法
Feature matching can be categorized into detector-based and detector-free methods. Previous research~\cite{LIFT, SuperPoint, SuperGlue, kunhong_cvpr2022} predominantly focuses on detector-based strategies, which first detect keypoints that are robust to significant appearance and viewpoint variations, thereby %avoiding additional computational costs and 
reducing potential matching noise caused by ambiguous features. 
After detection, feature description and all-pairs matching strategies are used to identify reliable correspondences.
{However, these multi-stage strategies heavily rely on the quality of detected keypoints and may fail when their quantity is insufficient, distinctiveness is ambiguous, or localization precision is compromised.} %detection is unsuccessful. 
Recent advancements bypass  point detection and perform dense, pixel-wise matching on low-resolution feature maps \cite{Sparse-NCNet, DRC-Net}. To further enhance accuracy and robustness, these methods focus on developing more robust feature descriptors~\cite{LoFTR,RoMa}, improving global matching strategies~\cite{DKM}, or incorporating fine-grained refinement techniques~\cite{RoMa}.
For instance, 
% loftr 主要在特征描述层面进行了 attn 增强
Sun \textit{et al.} \cite{LoFTR, LoFTR_PAMI} employed linear Transformers across views to augment low-resolution feature descriptions with global context, enhancing the robustness of the descriptors in textureless regions.
% RoMA 使用 LVM 来提高粗分辨率的特征鲁棒性
Edstedt \textit{et al.}~\cite{RoMa} replaced traditional CNNs with the frozen DINOv2~\cite{DINOv2}, yielding more robust coarse feature descriptions.
% 我们的做法
In line with these approaches, our method leverages large vision models to obtain robust feature representations across all scales.

% 另一个赛道, 使用光流 + 置信度估计的方法来得到特征匹配的结果
% Confidence-aware optical flow frameworks also facilitate dense matching.
{Unlike aforementioned methods that filter unreliable points before matching, recent research jointly estimates all-pairs correspondences with confidence, following post-matching filtering to discard unreliable pairs.}
% With estimated flow maps, 
Zhang \textit{et al.}~\cite{RGM} designed an uncertainty network to mask unreliable optical flow predictions in occluded or homogeneous regions.
% However, such a two-stage framework inevitably introduces additional computational costs.
% Some approaches unify flow and uncertainty prediction into a single forward pass.
Wannenwetsch \textit{et al.}~\cite{ProbFlow} converted energy functions into Gibbs distributions to jointly model optical flow and uncertainty.
Yin \textit{et al.}~\cite{HD3} treated the motion of each pixel as a discrete random variable, using a discrete probability distribution to calculate both optical flow and uncertainty.
Wang \textit{et al.}~\cite{SEA-RAFT} adopted mixture Laplacian parameterized distributions for probabilistic modeling.
Truong \textit{et al.}~\cite{PDC-Net,PDC-Net+} parameterized predictive distributions as constrained mixture models to jointly represent accurate predictions and outliers. Additionally, they introduced an uncertainty decoder to address overconfidence in {estimation results}. %estimates.
These methods demonstrate that confidence estimation can be integrated into optical flow frameworks for dense feature matching. 
% Inspired by these advances, our method incorporates confidence estimation into a unified framework, while decoupling the training of flow and confidence heads to mitigate mutual interference for stable optimization.
Inspired by these advances, we also employ a two-stage approach for feature matching, involving pixel-wise matching followed by filtering of unreliable estimates. 

\subsection{LVMs' Application on Correspondence Matching}
\label{sec:LVM_application}
Inspired by the rapid development of large vision models like CLIP~\cite{clip}, SAM~\cite{SAM} and DAM~\cite{depth_anything}, recent researchers try to expend their applications to downstream tasks through partial parameter fine-tuning~\cite{DenseAdaptor, LoRA-SAM}. 
However, such works mainly focus on the single-view tasks like {detection and segmentation}. 
Recently, Zhang \textit{et al.} \cite{FormerStereo} adopted frozen LVMs for domain generalized stereo matching. 
Liu \textit{et al.} \cite{ViTAStereo} also leveraged the frozen LVM representations as the input, and designed an elaborate cross-view enhance module to achieve task adaptation for stereo matching.
Bartolomei \textit{et al.} \cite{StereoAnywhere} proposed StereoAnywhere, leveraging the normal priors in DAMv2~\cite{DAMv2} to enhance the generalized capacity of RAFT-Stereo.  
Zhou \textit{et al.} \cite{SAMFlow} utilized the frozen SAM~\cite{SAM} backbone to enhance the context representations, which facilities the generalization ability to the fragmentation attack scenes.
Edstedt \textit{et al.} \cite{RoMa} treated the frozen DINOv2 as the robust but coarse feature matcher for dense feature matching.
% These methods treat the features from LVMs as additional information to enhance the feature representations and still highly rely on the source domain to pretrain the main feature backbone, while we regard these features are enough for dense matching to achieve robust and accurate estimations.
{Different from these methods that treat LVM features as auxiliary priors to enhance task-specific in-domain performance, we focus on unleashing the potential of LVM for cross-domain multi-task matching capabilities.}

%--------------------------------------------------------------------------
\section{Unified Matching Formulation}
In this section, we present our unified matching formulation. Specifically, 
we first discuss why different correspondence tasks can be unified as a single problem in Sec.~\ref{sec:unified_representation}. Then, we introduce how to derive task-specific geometric correspondences from the unified representation in Sec.~\ref{sec:conversion_mechanism}.

\subsection{Unified Representation}
\label{sec:unified_representation}
% insights and motivations
Intuitively, unifying two-frame correspondence tasks requires a unified representation to accommodate various types of geometric correspondences, such as disparity, flow, depth, and keypoint correspondences.
Our key insight is that all these tasks can be formulated as {all-pairs pixel matching} problems, {since they share a common objective of establishing geometric correspondences between pixels}. % with confidence estimation. %, thereby enabling addressing different tasks using a unified network architecture. 
{Specifically, stereo matching and unrectified stereo depth estimation can be viewed as 1D displacement estimation problems along epipolar lines. Optical flow estimation corresponds to a 2D displacement estimation problem, while feature matching involves an additional confidence filtering step to yield sparse but reliable 2D displacement estimation.}
These tasks differ only in the bias of displacement distribution and task-specific priors (\textit{e.g.}, epipolar constraints, photometric consistency assumption, and keypoint distinctiveness). 
From this point of view, these tasks can be formulated as the prior-free prediction of pixel-wise 2D displacement field $(\Delta u, \Delta v)$. Such a unified 2D displacement formulation eliminates the need for designing specialized unified architectures or
task-specific ensemble models, allowing for constructing a versatile foundation model for robust correspondence matching.
%and employing shared frameworks (\textit{e.g.}, feature extraction, correlation volume construction, and offset regression), we unify these tasks under a common network architecture and output format.}%, and task-specific outputs (e.g., disparity map and optical flow) can be obtained through parameter-free transforms.}
%\yj{Furthermore, with the unified 2D offset formulation, we can transform existing correspondence datasets ranging from stereo matching, optical flow, feature matching and depth estimation into the unified formulation to support large-scale pretraining.}

\subsection{Conversion Mechanism}
\label{sec:conversion_mechanism}
\begin{figure}[ht]
    \centering
    \includegraphics[width=0.35\textwidth]{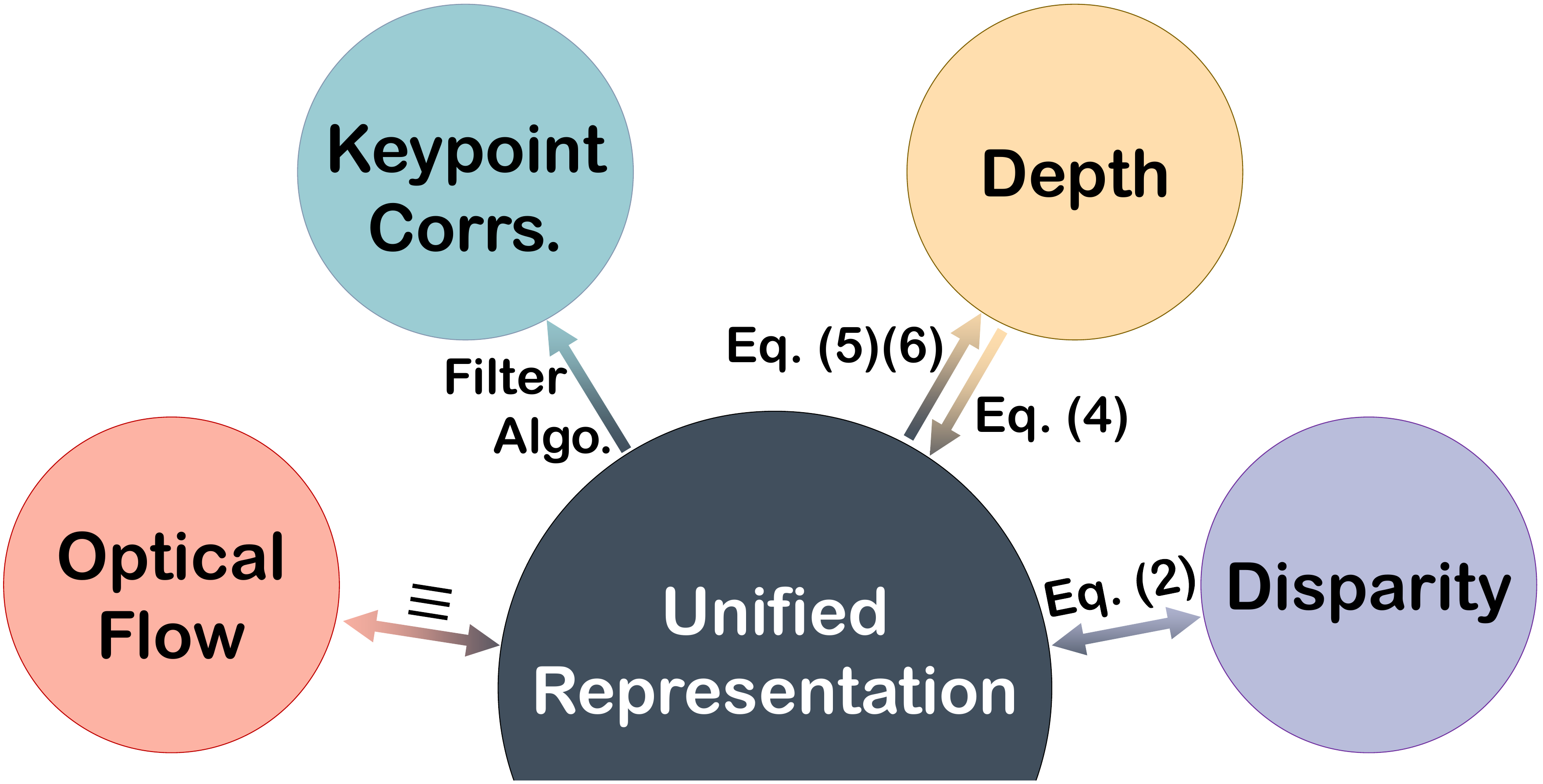}
    \caption{Conversion mechanism.}
    \label{fig:conversion}
\end{figure}
% 将 stereo, dense matching 转成 flow 的表达形式
Given a pair of images $\mathcal{I} = \{\mathbf{I}_{ref}, \mathbf{I}_{tar}\}$, {our unified formulation aims at finding the pixel-wise correspondences from the reference image to the target one. These correspondences are represented using pixel coordinate displacements as}:
\begin{equation}
    \mathbf{I}_{ref}(u_1, v_1) = \mathbf{I}_{tar}(u_2 - \Delta u, v_2 - \Delta v),
\end{equation}
where $(\Delta u, \Delta v)$ denotes the ground-truth displacement vector from the reference image $\mathbf{I}_{ref}(u_1, v_1)$ to the target one $\mathbf{I}_{tar}(u_2, v_2)$. 
{This representation is consistent with the standard definition of optical flow, therefore, the displacement field can directly serve as the output of optical flow task.
Similarly, keypoint correspondences required for feature matching can be derived from the dense displacement field through post-processing, such as applying forward-backward cyclic consistency to extract correspondences within the overlapping view region.}

Moreover, if the image pair is a rectified stereo pair, with the reference image corresponding to the left view, the disparity can be derived from the displacement as:
\begin{equation}
    \Delta D = - \Delta u, \Delta v = 0,
    \label{eq:flow2disparity}
\end{equation}
since disparity is defined as
\begin{equation}
    \mathbf{I}_{ref}(u_1, v_1) = \mathbf{I}_{tar}(u_2 + \Delta D, v_2),
\end{equation}
where $\Delta D$ denotes the ground-truth disparity.
Additionally, if the pair is captured in a regular static scene without moving objects, the scene depth can be inferred using the known camera pose $[\mathbf{R} \ \mathbf{T}]$ and intrinsics $\mathbf{K}$ according to the pinhole camera projection model:
\begin{equation}
    s_2 \begin{bmatrix} u_2\\ v_2\\ 1\\ \end{bmatrix} = 
    \mathbf{K}_2 (\mathbf{R}_2 \mathbf{R}_1^{-1} (\mathbf{K}_1^{-1}\begin{bmatrix} u_1\\ v_1\\ 1\\ \end{bmatrix}Z -\mathbf{T}_1) + \mathbf{T}_2),
    \label{eq:pinhole_camera_model}
\end{equation}
where $u_2 = u_1 + \Delta u$, $v_2 = v_1 + \Delta v$. 
By defining $\mathbf{H}=\mathbf{K}_2 \mathbf{R}_2 \mathbf{R}_1^{-1} \mathbf{K}_1^{-1}$, $\mathbf{B}=-\mathbf{K}_2 \mathbf{R}_2 \mathbf{R}_1^{-1} \mathbf{T}_1 + \mathbf{K}_2 \mathbf{T}_2$, we simplify the matrix and can infer the pixel-wise depth as 
\begin{equation}
\resizebox{0.88\hsize}{!}{$
    Z_u(u_1,v_1) = \frac{B_{31} u_2 - B_{11}}{(H_{11} u_1 + H_{12} v_1 + H_{13}) - (H_{31} u_1 + H_{32} v_1 + H_{33}) u_2 }
$},
\label{eq:flow2depth_method1}
\end{equation} 
or 
\begin{equation}
\resizebox{0.88\hsize}{!}{$
    Z_v(u_1,v_1) = \frac{B_{31} v_2 - B_{21}}{(H_{21} u_1 + H_{22} v_1 + H_{23}) - (H_{31} u_1 + H_{32} v_1 + H_{33}) v_2 }
$}.
\label{eq:flow2depth_method2}
\end{equation} 
% 根据光流分量的大小来选择具体应用哪一个计算方式, 越大的光流, 计算深度的误差越小
Theoretically, the depths calculated using Eq.~\ref{eq:flow2depth_method1} or Eq.~\ref{eq:flow2depth_method2} should be consistent. 
% However, in practice, due to \yj{the calibration errors in camera internal and external parameters}, %numerical computation errors and issues such as division by zero, 
However, they may not align in practice due to {the matching noise and calibration errors.}
% the depth values obtained from the two methods may not align. 
% To address this inconsistency, we mask regions where the difference in depth values exceeds 0.3 meters, and use the average of two calculated depth map as the final depth estimation results.
% We select either Eq.~\ref{eq:flow2depth_method1} or Eq.~\ref{eq:flow2depth_method2} according to the numerous scale of $\Delta u$ and $\Delta v$ to avoid the numerical calculation errors caused by zero division. That is, choose Eq.~\ref{eq:flow2depth_method1} if $\Delta u \ge \Delta v$ otherwise Eq.~\ref{eq:flow2depth_method2}.
To address this inconsistency, we optimize $Z(u_1, v_1)$ using least squares as $Z_{lsm}(u_1, v_1)=(\mathbf{A}^\mathrm{T} \mathbf{A})^{-1} \mathbf{A}^\mathrm{T} \mathbf{b}$, where
\begin{equation}
\resizebox{0.88\hsize}{!}{$
    \mathbf{A} = \begin{bmatrix}
        {(H_{11} u_1 + H_{12} v_1 + H_{13}) - (H_{31} u_1 + H_{32} v_1 + H_{33}) u_2 } \\
        {(H_{21} u_1 + H_{22} v_1 + H_{23}) - (H_{31} u_1 + H_{32} v_1 + H_{33}) v_2 }
    \end{bmatrix}
$},
\label{eq:depth_lsm}
\end{equation}
\begin{equation}
    \mathbf{b} = \begin{bmatrix}
        {B_{31} u_2 - B_{11}} \\
        {B_{31} v_2 - B_{21}}
    \end{bmatrix}.
\label{eq:depth_lsm}
\end{equation}

%-------------------------------------------------------------------------------
\begin{figure*}[ht]
    \centering
    \includegraphics[width=0.9\textwidth]{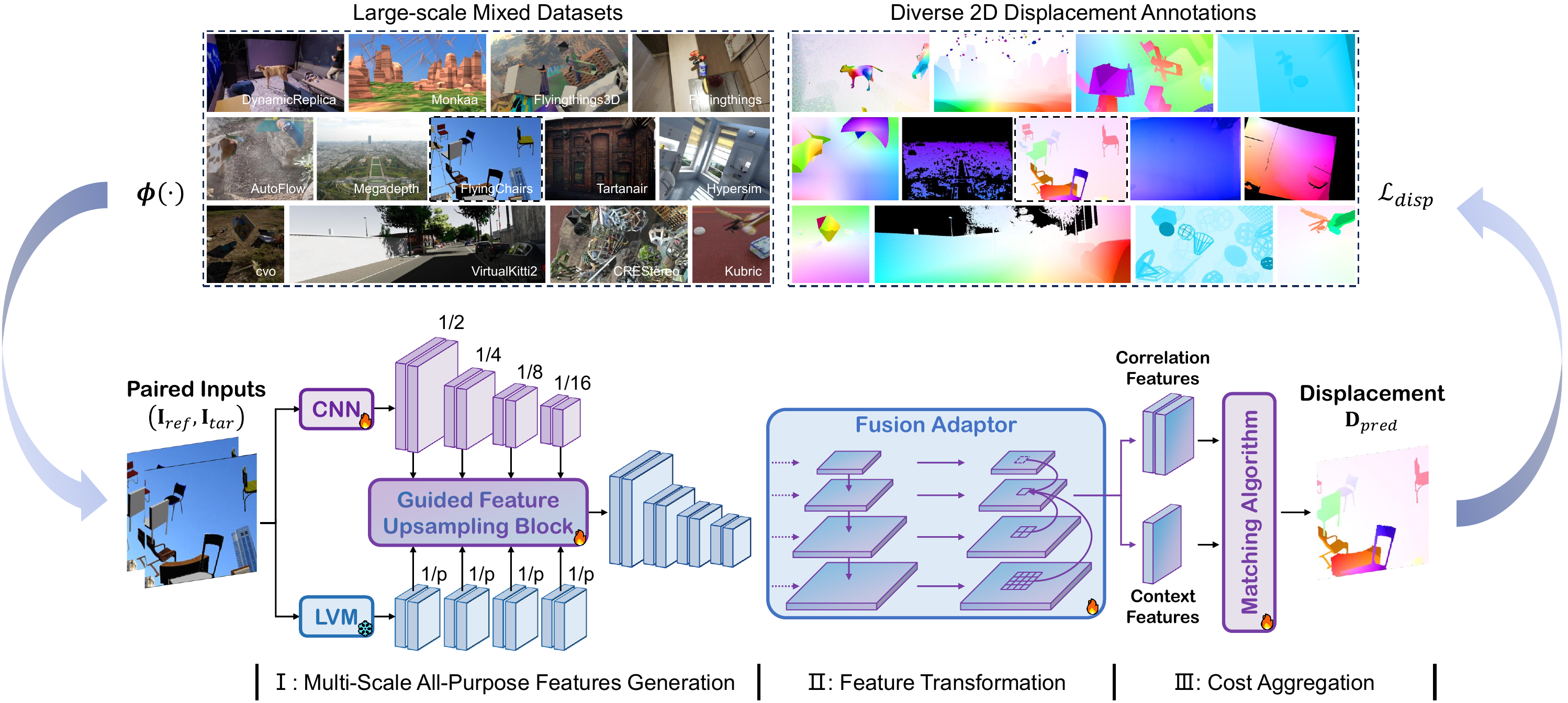}
    \caption{An overview of the proposed framework. %Our framework can be seamlessly integrated with existing architectures by employing their feature backbones as CNN feature extractor. 
    $\phi(\cdot)$ means data augmentation strategies described in Sec.~\ref{sec:exp_development}.}
    \label{architecture}
\end{figure*}
\section{Unified Matching Model}
% 统一不同任务后, 需要做什么? 为什么原来的方法不行?
% To enhance cross-domain generalization, we highlight the importance of robust feature representations, and use all-purpose features as the solution.
% With the aforementioned unified formulation, the core challenge of cross-task generalization shifts toward cross-domain generalization. Inspired by \cite{FormerStereo}, we highlight the importance of robust feature representations, and use all-purpose features as the solution.
% Motivated by the impressive zero-shot performance of LVMs on downstream applications, we treat the LVM as a feasible solution for constructing robust feature extractor.
{With the aforementioned unified formulation, as well as its representational equivalence to optical flow, we develop a unified model based on existing optical flow baselines. However, the limited feature representation capacity of these baselines compromises cross-task and cross-domain generalization.
Motivated by the impressive zero-shot performance of LVMs on diverse downstream applications~\cite{FormerStereo, SAMFlow, RoMa}, we leverage LVMs to build a general and robust feature extractor to achieve generalization gains.}
In particular, we regard all-purpose features produced by ViT-based foundation models~\cite{DINOv2, SAM,depth_anything} as domain-invariant representations, and adapt such features for the dense correspondence task.
%This formulation requires retrieving fine-grained and matching-oriented representations from all-purpose features. To this end, we propose a guided feature upsampling block to recover fine-grained details in the feature map. Then, a general adapter is constructed to transfer the resultant features for matching.

% overview
We provide an overview of our proposed framework in Fig.~\ref{architecture}, which can be separated into three parts: multi-scale all-purpose feature generation, feature transformation, and cost aggregation. 
Specifically, given a pair of images $\mathcal{I}=\{\mathbf{I}_{ref}, \mathbf{I}_{tar}\}$, we first obtain the multi-scale features $\{\mathbf{G}_{1/2}, \mathbf{G}_{1/4}, \mathbf{G}_{1/8}, \mathbf{G}_{1/16}\}$ from CNN and multi-layer all-purposed features $\{\mathbf{F}_{l_1}, \mathbf{F}_{l_2}, \mathbf{F}_{l_3}, \mathbf{F}_{l_4}\}$ from ViT-based foundation model, respectively. 
{The CNN features guide the interpolation of the plain ViT features into pyramid ones via the guided upsampling module.
These pyramid features are subsequently processed by a feature pyramid network (FPN) for multi-scale integration and task transfer.}
% Next, we restore the high-resolution all-purpose features using the guided upsampleing module, and then feed these features to the feature pyramid network for fusion and task transfer. 
Finally, the resultant multi-scale feature pyramid is adapted to a specific matching baseline for displacement estimation.

% organization
In the following, {we will first demonstrate our motivation that LVMs are superior to encoders trained on small-scale, task-specific datasets for multi-domain generalization in Sec.~\ref{sec:toy_example}.} Then, we
{introduce our feature transformation strategy for adapting LVMs to matching models, including the module design and loss constrains in Sec.~\ref{sec:LVM_adaptor}. 
We finally present the loss function in Sec.~\ref{sec:loss_function}.}

%------------------------------------------------------------------------------------
\begin{figure*}[ht]
    \centering
    \includegraphics[width=1.0\textwidth]{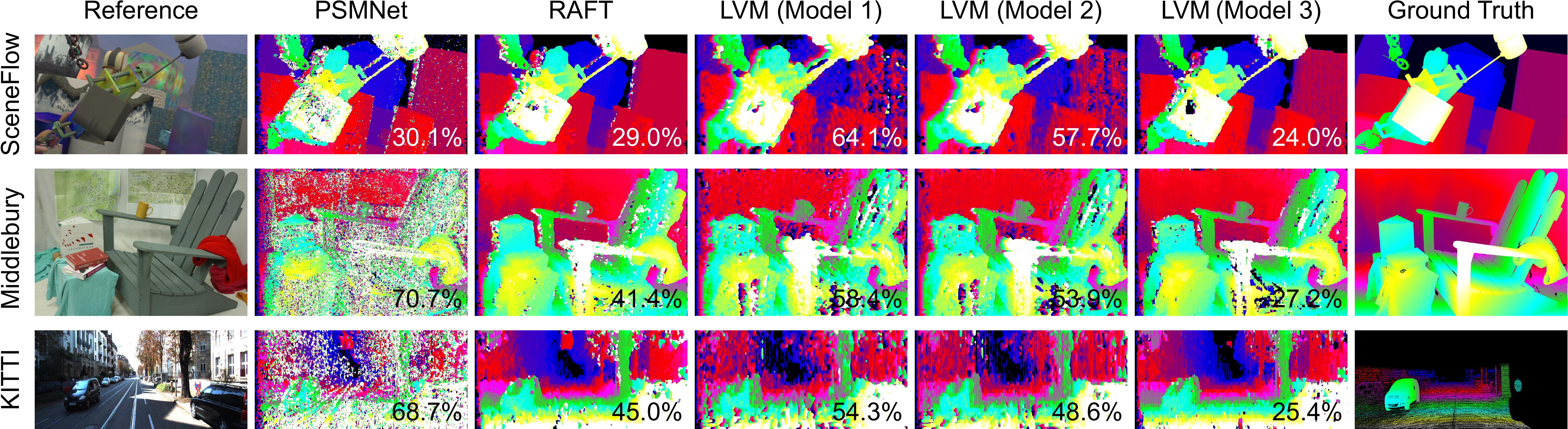}
    \caption{Disparity estimation from feature matching without cost aggregation. Features extracted by PSMNet~\cite{PSMNet}, RAFT~\cite{RAFT}, and DAMv2-small~\cite{DAMv2} are adopted for evaluation. All learnable strategies are only pretrained on the SceneFlow~\cite{DispNetC} dataset. The metric of Bad 3.0 is reported in each estimation.}
    \label{fig:toy_example}
\end{figure*}

\subsection{Motivation}
\label{sec:toy_example}

% Please add the following required packages to your document preamble:
% \usepackage{multirow}
\begin{table}[]
    \setlength{\tabcolsep}{2pt}
    \centering
    \caption{Toy Examples to validate the potential of all-purpose features for correspondence matching. We adopt DAMv2-small as the LVM encoder, comparing its feature alignment performence with feature extraction of those in stereo or flow estimation algorithms. The max considered disparity is 192 and the metric is Bad 3.0 (\%).}
    \scalebox{0.9}{
        \begin{tabular}{clcccc}
            \hline
            ID & \textbf{Model}              & \textbf{Scale}       & \textbf{FlyingThings}& \textbf{Middlebury}  & \textbf{KITTI 15}    \\ \hline
            0  & RAFT~\cite{RAFT}            & \multirow{3}{*}{1/8} & 61.9                 & 67.6                 & 58.3                 \\
            1  & LVM (single-layer)          &                      & 73.8                 & 79.6                 & 71.8                 \\
            2  & LVM (multi-layers)          &                      & 65.9                 & 72.5                 & 61.1                 \\ \hline
            % LVM (multi-layers) + Linear &                      & xx.x                 & xx.x                 & xx.x               \\ \hline
            0  & PSMNet~\cite{PSMNet}        & \multirow{5}{*}{1/4} & \underline{32.6}     & 66.4                 & 71.0                 \\
            0  & RAFT~\cite{RAFT}            &                      & 32.7                 & \underline{39.1}     & \underline{36.9}     \\
            1  & LVM (single-layer)          &                      & 60.8                 & 67.9                 & 57.6                 \\
            2  & LVM (multi-layers)          &                      & 53.9                 & 62.8                 & 49.0                 \\
            3  & LVM (multi-layers) + Linear &                      & \textbf{28.2}        & \textbf{39.0}        & \textbf{24.8}        \\ \hline
            % LVM (multi-layers) + Ours   &                      & \multicolumn{1}{l}{} & \multicolumn{1}{l}{} & \multicolumn{1}{l}{} \\ \hline
        \end{tabular}
    }
    \label{tab:toy_example}
\end{table}

% 在这里使用非参数化的方法来构造整体模型, 并逐一说明存在的问题, 以及我们的具体改进模块
We first investigate whether all-purpose features offer advantages over task-specific features in cross-domain matching. 
To simplify the analysis, we select 1D displacement estimation (\textit{i.e.}, stereo matching) as a case study and use DAMv2~\cite{DAMv2} as the LVM encoder, evaluating its performance in multi-domain scenarios. 
Specifically, given a stereo image pair, we first extract their all-purpose features using LVM, {obtaining $HW/p^2$ image tokens. Here, $p$ refers to the patch size, $H$ and $W$ denote the height and width of the image, respectively. Next, the resultant image tokens are reshaped to obtain patch-level all-purpose features $\mathbf{F}_{l_i} \in \mathbb{R}^{D \times \frac{H}{p} \times \frac{W}{p}}$ with $D$ channels, which undergo bilinear interpolation to achieve high spatial resolution, and then be employed to construct a 3D cost volume using cosine similarity across all disparity levels.} Finally, we apply trilinear interpolation to the cost volume to match the input resolution and compute disparity map via $\mathrm{argmax}(\cdot)$ along the disparity dimension (denoted as Model 1). For fair comparison, we implement a baseline (Model 0) using features from either PSMNet~\cite{PSMNet} or RAFT~\cite{RAFT} encoders processed through identical steps without cost aggregation.

{To leverage diverse semantic representations}, we divide LVM into four segments and collect output tokens of each segment to form multi-layer features. These features are employed to construct four distinct cost volumes, which are then fused through dot product (Model 2). 
Furthermore, we replace bilinear interpolation with deconvolution~\cite{deconvolution} layers, pretraining this variant on FlyingThings~\cite{FlowNet} for matching-oriented adaptation to improve performance (Model 3).
% We also provide a variant of FormerStereo, denoted as Former, which is pretrained on SceneFlow without cost aggregation, to validate the effectiveness of the proposed Feature Transformation module. Note that the Former acts as a feature transformer to transfer all-purpose features to matching-specific ones.
% 实验结果

As quantified in Table~\ref{tab:toy_example}, all-purpose features demonstrate consistent cross-domain performance. 
Besides, increasing scales of matching features and utilizing multi-layer features further improve matching accuracy, outperforming the poorly generalized PSMNet on unseen real-world scenes. 
These observations indicate that comprehensive utilization of all-purpose features provides significant advantages in domain generalization compared to task-specific representations.
Furthermore, LVMs combined with simple linear adaptation not only maintain cross-domain consistency but also surpass the well-generalized RAFT in both seen and unseen domains.  
These findings highlight the potential of all-purpose features for zero-shot matching.
Qualitative comparisons in Fig.~\ref{fig:toy_example} provide intuitive confirmation that all-purpose features contribute less noisy disparity across domains.
Based on these observations, we subsequently employ LVMs as domain-invariant encoders and unleash their potential for matching tasks in the following section.

\subsection{General-to-Matching Feature Transformation}
\label{sec:LVM_adaptor}
% 在 toy example 的基础上, 还有哪些改进空间? 如何进一步释放 all-purpose features 的潜能? 
% 包括: 更有效的特征上采样方案, 多尺度信息聚合、跨视角约束
% \yj{We propose a general feature transformation pipeline that leverages the domain-invariant properties of LVM features to enhance the generalization performance of the selected baseline method.
% To achieve this goal, we design three core components:
{We aim to leverage the domain-invariant properties of LVM features to enhance the generalization performance of any selected baseline method. However, directly integrating an LVM with a cost aggregation module produces inferior performance due to mismatches in feature scale, dimension, and pretraining objectives. 
To address these challenges, we propose a general feature transformation pipeline comprising three core components:
(\romannumeral 1) a \emph{position-aware guided feature upsampling block} that adaptively interpolates all-purpose features for fine-grained matching while preserving their domain-invariant characteristics;
(\romannumeral 2) a \emph{hierarchical adaptation network} that supports multi-layer feature fusion and matching-oriented feature extraction, while accommodating various input formats of cascaded matching baselines;
(\romannumeral 3) an explicit \emph{cross-view matching constraint} mechanism that enhances matching-specific attributes within all-purpose features to facilitate task adaptation.}
\subsubsection{Position-Aware Guided Feature Upsampling}
\begin{figure}[t]
    \centering
    \includegraphics[width=0.45\textwidth]{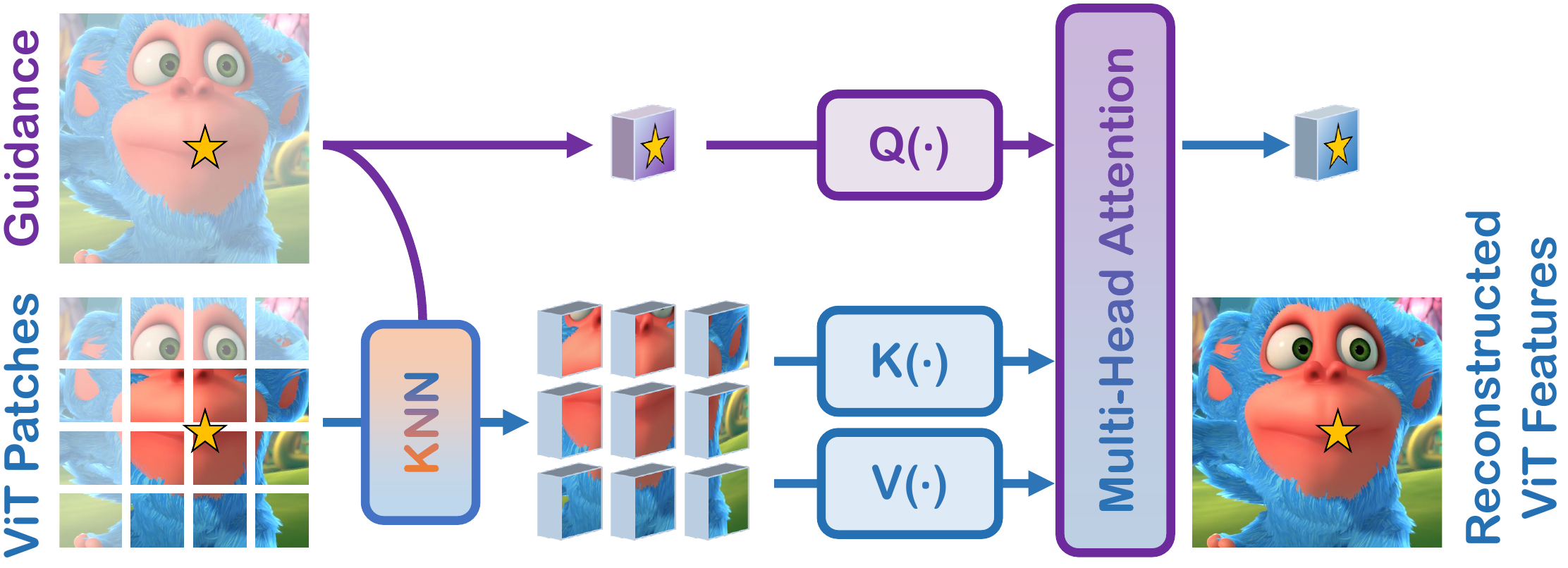}
    \caption{An overview of the guided feature upsampling block.}
    \label{fig:guided_upsampling}
\end{figure}
% 单独使用 LVM 的特征, 有什么优势, 存在什么问题
Features extracted by LVMs are typically low-resolution, making them insufficient for fine-grained matching. 
% 针对这些问题, 在不引入额外的编码特征的前提下, 有什么可行的方案
% A na\"ive solution is to apply interpolation or learnable methods to upscale the feature maps.
{A na\"ive solution is to apply interpolation to upscale the feature maps.
However, existing upsampling strategies typically use fixed, position-independent convolutional kernels ${\omega}$ to obtain $s$-times high-resolution outputs $\mathbf{F}^{h}$ from low-resolution inputs $\mathbf{F}^{l}$ as:}
\begin{equation}
    \mathbf{F}^{h}(u,v) = \sum_{i,j} \omega(i,j) \cdot \mathbf{F}^{l}\left(\frac{u}{s}+i,\frac{v}{s}+j\right),
\end{equation}
which inherently limits their contextual adaptability and cross-domain generalization. 
{Instead of directly regressing high-resolution details through static kernels, we propose a position-aware guided feature upsampling strategy, which dynamically adjusts the upsampling kernel weights at different spatial positions during inference.}
Specifically, we first leverage a lightweight {FPN encoder} to generate target-resolution guided features, which are treated as the structure-preserving guidance.
% We then employ a cross-attention mechanism~\cite{Transformer} to align CNN-based and ViT-based features based on content similarity, yielding the desired convex combination weights for feature recomposition.
% To associate the CNN-based and ViT-based features, we leverage the attention mechanism to generate the convex weights with feature similarity. 
% More specifically, 
Then, given an arbitrary position $\mathbf{p}=(u_0,v_0)$ in guidance, as marked by a star in Fig.~\ref{fig:guided_upsampling}, we project its features into the query space, denoted as $\mathbf{F}_{Q}^{h}\in\mathbb{R}^{H \times W \times C}$. 
Meanwhile, all-purpose features of the correspondent k-nearest neighbors are projected into the key and value spaces, denoted as $\mathbf{F}_{K}^{l}\in\mathbb{R}^{h \times w \times C}$ and $\mathbf{F}_{V}^{l}\in\mathbb{R}^{h \times w \times C}$, respectively. Here, $H \times W$ represents the target resolution for interpolation, and $h \times w$ means the current size of all-purpose features. %, and $N=9$ indicates the 3$\times$3 nearest neighbors $q \in \mathcal{N}(p)$ in all-purpose features.
% Given a target position $p$ in guided features and its correspondence 3x3 nearest neighbors $q \in \mathcal{N}(p)$ in all-purpose features, the output features at position $p$ is formulated as:
Finally, the high-resolution features $\mathbf{F}^{h}$ at position $p$ are recomposed as:
\begin{equation}
    \mathbf{F}^{h}(\mathbf{p}) = \sum_{\mathbf{q} \in \mathcal{N}(\mathbf{p})} \omega(\mathbf{p},\mathbf{q}) \cdot \mathbf{F}_{V}^{l}(\mathbf{q}),
\end{equation}
where $\mathbf{q} \in \mathcal{N}(\mathbf{p})$ indicates the 3$\times$3 nearest neighbors of $p$ in all-purpose features, and $\omega(\mathbf{p},\mathbf{q})$ denotes the normalized feature similarity scores between query features of pixel $\mathbf{p}$ and key features of pixel $\mathbf{q}$, \textit{i.e.},
\begin{equation}
    \omega(\mathbf{p},\mathbf{q}) = \frac{\exp\left(\mathbf{F}_{Q}^{h}(\mathbf{p}) \cdot \mathbf{F}_{K}^{l}(\mathbf{q}) / \sqrt{C}\right)}{\sum_{\mathbf{q}^* \in \mathcal{N}(\mathbf{p})} \exp\left(\mathbf{F}_{Q}^{h}(\mathbf{p}) \cdot \mathbf{F}_{K}^{l}(\mathbf{q}^*) / \sqrt{C}\right)}.
    \label{eq:convex_combination_weights}
\end{equation}
Note that in Eq.~\ref{eq:convex_combination_weights}, higher similarity scores contribute larger convex combination weights, ensuring that the upsampling process restores sharp edges, contours, and even thin structures, in accordance with the guided features.

Given the similarity between the proposed guided upsampling strategy and the attention mechanism~\cite{Transformer}, we implement it using an attention-based design and further improve its performance with multi-head attention.  
To accelerate weight generation through matrix multiplication, we unfold the key and value features using a $3\times3$ kernel and apply nearest-neighbor interpolation to obtain high-resolution but coarse representations, denoted as $\mathbf{F}_{K}^{h}, \mathbf{F}_{V}^{h} \in \mathbb{R}^{nHW \times 9 \times (C/n)}$, where $n$ means the number of attention heads. 
The query features are reshaped as $\mathbf{F}_{Q}^{h} \in \mathbb{R}^{nHW \times 1 \times (C/n)}$, and then utilized to construct the upscaled features as:
\begin{equation}
    \mathbf{F}^{h} = \mathrm{Softmax}\left(\mathbf{F}_{Q}^{h} \times \left(\mathbf{F}_{K}^{h}\right)^\mathrm{T} / \sqrt{C/n}\right) \times \mathbf{F}_{V}^{h}.
\end{equation}
Finally, we reshape the features size from $\mathbb{R}^{nHW \times 1 \times (C/n)}$ to $\mathbb{R}^{H \times W \times C}$, and apply an MLP for feature projection.

\subsubsection{Hierarchical Adaptation Network}
\textbf{Multi-scale Feature Fusion}. 
% 到目前为之仍未解决的问题
Although domain bias has been mitigated by the use of all-purpose features and scale-related issues have been addressed through guided upsampling, simply grafting matching baselines with LVMs remains suboptimal.
% 为什么要引入多尺度解码器? 在编码与特征引导之后还存在什么问题影响模型性能? 引入多尺度有什么好处?
% 采用 decoder 的两个用途: (1) 多层级特征融合, 解决单层级特征在匹配任务上的表征局限性; (2) 提供了一个从 all-purposed token 中滤除无关信息, 使表征更关注跨视角匹配的滤波器
{This limitation arises because LVMs are not pretrained for correspondence matching.
As a result, the all-purpose features may not align with the task of accurate correspondence matching.}
%Thus, extracting task-specific information for correspondence matching remains essential.
%Furthermore, as demonstrated in Table~\ref{tab:toy_example}, incorporating multi-layer features improves the generalization capabilities of LVM-based feature alignment. \yj{However, this strategy is not yet adopted in the current design.}
% We assume that all-purpose features from different levels of an LVM share the same spatial scale but differ in semantic content and feature emphasis, making them complementary. 
% This observation emphasizes the significance of multi-layer feature fusion.

To remedy the aforementioned issue, we propose a U-shaped fusion adapter. %network incorporating guided upsampling modules. 
Specifically, we first employ a multi-stage CNN encoder to generate multi-scale guidance features. %, which are then utilized to rescale all-purpose features at multiple levels through position-aware guided upsampling blocks. 
Meanwhile, we collect multi-layer all-purpose features from several segments of LVM, establishing a one-to-one correspondence with the multi-scale CNN features.
% for upsampling, yielding an all-purpose feature pyramid. 
Next, we construct multi-scale all-purpose features via guided upsampling blocks, wherein each scale-specific feature map is recomposed by upsampling the corresponding all-purpose features under the guidance of target-scale CNN features.
We then employ ConvNeXt blocks~\cite{ConvNeXt} to construct the standard feature pyramid decoder, %progressively integrating high-level semantic information from deeper, low-resolution features into shallower, high-resolution features, thereby producing the required feature pyramid. 
% In practice, our CNN encoder consists of four stages, each producing feature maps at resolutions of 1/2, 1/4, 1/8, and 1/16 relative to the original image size. Similarly, the decoder outputs features across four stages to ensure compatibility with input requirements for most matching baselines.
resulting in feature pyramid at resolutions of 1/16 1/8, 1/4, and 1/2 relative to the original image size.

% Different to U-net adopting the feature pyramid network to enlarge the receptive field, Transformer has a global receptive field in all level. However, the shallow layer features have limit representation capacity. Even though we use guide upsampling to upscale its size, the single-level features can't not xxx.
% To retrieve fine-grained details from the all-purposed tokens, we introduce the multi-layer image reconstruction task as the auxiliary task. To adapt the features while maximizing the preservation of domain-invariant representations in all-purpose features, we introduce the feature similarity constrain on the final fusion level.

\textbf{Multi-scale Patch Embedding}.
{U-Net progressively aggregates multi-scale features ranging from 1/16 to 1/2 resolution. However, most methods build cost volumes only at the 1/8 scale, neglecting the richer details captured at 1/4 and 1/2 resolutions.}
To address this limitation, we propose a multi-scale patch embedding block to effectively fuse features from multiple scales into a unified target feature space at one step.
Specifically, we use patch sizes of 4, 2, 1, and 1/2 to fold the feature maps at scales of 1/2, 1/4, 1/8, and 1/16, respectively. 
This design ensures that all scales features are position-aligned with the 1/8 scale one. 
Subsequently, we apply four independent linear projections to compress and transform features at each scale, which are then concatenated along the feature dimension. Finally, a multi-layer perceptron (MLP) is used to generate the fused features at the target scale. %Our approach fully exploits multi-scale and multi-layer information, thus enhancing feature representation capacity at the target scale. 

\subsubsection{Cross-View Matching Constraint}
\begin{figure}[t]
    \centering
    \includegraphics[width=0.48\textwidth]{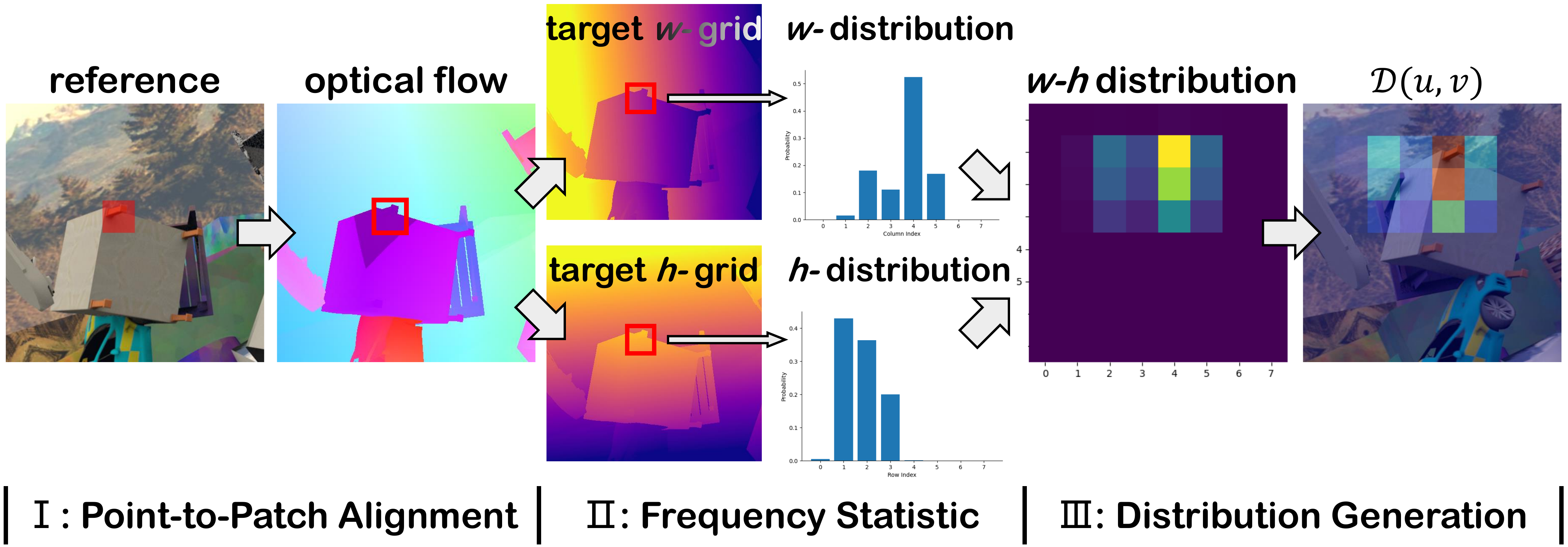}
    \caption{A pipeline to obtain the patch correspondence distribution from ground-truth optical flow map.}
    \label{fig:illustrate_infoNCE}
\end{figure}
% 将通用特征向跨视角匹配特征进行迁移
% 必要性, 在进行参数化的变换和映射时会损失跨视角一致性的表征
The parameterized feature transformation and projection only account for the single-view constraints, {overlooking the extraction of cross-view consistent representations.}
% 视差约束很难直接约束视差代价体的分布
Although the displacement supervision~\cite{FlowNet, RAFT} provides implicit guidance for aligning cross-view representations at matching points, it places only weak constraints on unrelated proposals, raising the risks of multi-modal distribution and ultimately degrades the quality of the regression results, as demonstrated in Fig.~\ref{fig:ablation_infoNCE}. 
% Therefore, it is necessary to purify the matching-specific features from fine-grained representations at first.
% 我们的做法: 约束和匹配二合一, 保留对视差估计有益的语义和匹配信息
To emphasize the importance of learning cross-view consistent representations, we leverage contrastive learning for enhancing the feature similarity on the correspondences while impress on the other regions. 
% 余弦相似度的优势
% The cosine metric calculates the pure similarity information of cross-views, so that avoids the negative impacts of short-cut information to matching performance. 
% concate volume 的优势
% Meanwhile, the concatenation operation preserves the semantic information, encouraging context reasoning in occluded and textureless regions during cost aggregation.
% The proposed C4 space incorporate the generality of cosine similarity into the feature concatenation based cost volume, so that benefit from both.
% The C4 space utilizes normalized features constrained by cosine similarity to construct concatenation cost volume, implicitly encoding the matching prior into the initial cost volume.

% 具体技术细节
Specifically, given paired features $\{\mathbf{F}_{ref}, \mathbf{F}_{tar}\}$, we first normalize these representations along feature dimension and then calculate the cosine similarity of normalized feature pairs under each flow proposal to obtain the 3D score volume $\mathbf{S} \in \mathbb{R}^{h \times w \times hw}$ as:
\begin{equation}
    \mathbf{S}(u,v,\mathbf{f}) = \sum_c \mathbf{F}_{ref}(u,v) \cdot \mathbf{F}_{tar}(u+\mathbf{f}_{u}, v+\mathbf{f}_{v}).
    \label{score_volume}
\end{equation}
To ensure that the score volume is provided with similarity prior, we apply a pixel-wise infoNCE~\cite{infoNCE} loss to the 3D score volume, increasing the proportion of matching pixel score among all candidates so that easier to distinguish correspondences among proposals. This is formulated as:
\begin{equation}
\resizebox{0.9\hsize}{!}{$
    \mathcal{L}_{NCE}(u,v) = \sum_{\mathbf{f}} - p(u,v,\mathbf{f}) \cdot \log \frac{\exp \left( \mathbf{S}(u, v, \mathbf{f})/\tau \right)}{\sum_{\mathbf{f}^{'}} \exp \left( \mathbf{S}(u, v, \mathbf{f}^{'}) / \tau \right)} 
$},
    \label{infoNCE_loss}
\end{equation}
{where $p(u,v,\mathbf{f})$ indicates the probability of proposal $\mathbf{f}$ in the ground-truth discrete distribution $\mathcal{D}(u,v)$}, $\tau$ is a temperature coefficient and we set $\tau=0.07$ following~\cite{FCStereo}. The ground-truth discrete distribution $\mathcal{D}(u,v)$ is derived from ground-truth optical flow map, as illustrated in Fig.~\ref{fig:illustrate_infoNCE}.
Specifically, we first compute the target coordinates within the reference coordinate system using the ground-truth optical flow.
Due to the resolution discrepancy between the feature map and the coordinate map, each feature point corresponds to a small patch on the coordinate maps. We therefore quantize the coordinates within each patch to obtain frequency counts along both horizontal and vertical axes.
Finally, the ground-truth flow distribution is generated from the joint distribution of these horizontal and vertical frequencies.
% , and the non-occluded mask is obtained from ground-truth disparity map according to the left-right consistency and uniqueness.

Note that we set $\mathcal{L}_{NCE}=\mathrm{mean}\left(\sum_{u,v} \mathcal{L}_{NCE}\left(u,v\right)\right)$ and intentionally do not exclude occluded regions, even though cosine similarity may not yield meaningful metrics there, as we aim to ensure that the output features exhibit global representational capabilities and respond appropriately to occluded areas.

\subsection{Loss Function}
\label{sec:loss_function}
We supervise our network using loss $\mathcal{L}_{disp}$, which follows the same formulation as the adopted baseline and measures the distance between the predicted and ground-truth displacements over the full sequence of predictions. In addition, we incorporate a contrastive learning loss $\mathcal{L}_{NCE}$ to promote the learning of cross-view consistent representations in the adopter module. The complete loss function is defined as
\begin{equation}
    \mathcal{L} = \mathcal{L}_{disp} + \mathcal{L}_{NCE}.
\end{equation}

\begin{table}[]
    \setlength{\tabcolsep}{2pt}
    \centering
    \caption{Datasets used for training PanMatch. Datasets used for fine-tuning are highlighted in boldface.}
    \scalebox{1.0}{
    \begin{tabular}{lcccc}
    \hline
    Dataset                                 & Annotation                                                              & Scene Type & Frames     & Image Size \\ \hline
    FlyingChairs~\cite{FlowNet}             & \multirow{10}{*}{\begin{tabular}[c]{@{}c@{}}Optical\\ Flow\end{tabular}}& Synthetic  & $\sim$23k  & 384x512    \\
    \textbf{FlyingThings}~\cite{DispNetC}   &                                                                         & Synthetic  & $\sim$81k  & 540x960    \\
    \textbf{Monkaa}~\cite{DispNetC}         &                                                                         & Outdoors   & $\sim$35k  & 540x960    \\
    AutoFlow~\cite{autoflow}                &                                                                         & Synthetic  & $\sim$27k  & 448x576    \\
    Dynamic Replica~\cite{dynamicstereo}    &                                                                         & Indoors    & $\sim$144k & 720x1280   \\
    Tartanair~\cite{tartanair}              &                                                                         & Outdoors   & $\sim$306k & 480x640    \\
    Kubric~\cite{kubric}                    &                                                                         & Synthetic  & $\sim$132k & 512x512    \\
    CVO~\cite{AccFlow}                      &                                                                         & Synthetic  & $\sim$125k & 512x512    \\
    \textbf{VirtualKITTI2}~\cite{vkitti2}   &                                                                         & Driving    & $\sim$42k  & 375x1242   \\ \hline
    \textbf{Hypersim}~\cite{hypersim}       & \multirow{2}{*}{Depth}                                                  & Indoors    & $\sim$367k & 768x1024   \\
    MegaDepth~\cite{megadepth}              &                                                                         & Outdoors   & $\sim$285k & 480x640    \\ \hline
    \textbf{CREStereo}~\cite{crestereo}     & \multirow{2}{*}{Disparity}                                              & Synthetic  & $\sim$200k & 1080x1920  \\
    FallingThings~\cite{fallingthings}      &                                                                         & Indoors    & $\sim$62k  & 540x960    \\ \hline
    \end{tabular}}
    \label{tab:datasets}
\end{table}

\section{Experiments}
In this section, we first describe the development of PanMatch in Secs.~\ref{sec:exp_development}, then compare it with both unified correspondence methods and task-specific methods in Sec.~\ref{sce:comparison_with_versatiles} and Sec.~\ref{sce:comparison_with_specifics} respectively, highlighting its versatile performance across multiple benchmarks. 
Next, we conduct an ablation study to assess our contributions in Sec.~\ref{sec:model_analyses}. 
% Additionally, we validate the plug-and-play properties of the proposed framework, as well as the performance improvement compared to our previous FormerStereo in Sec.~\ref{sec:play_and_plug}.
{Finally, we present the advantages of our unified models over independent methods for real-world application in Sec.~\ref{sec:application}.}

\subsection{Implementation Details}
\label{sec:exp_development}
\subsubsection{Training Datasets}
\begin{figure}[t]
    \centering
    \includegraphics[width=0.49\textwidth]{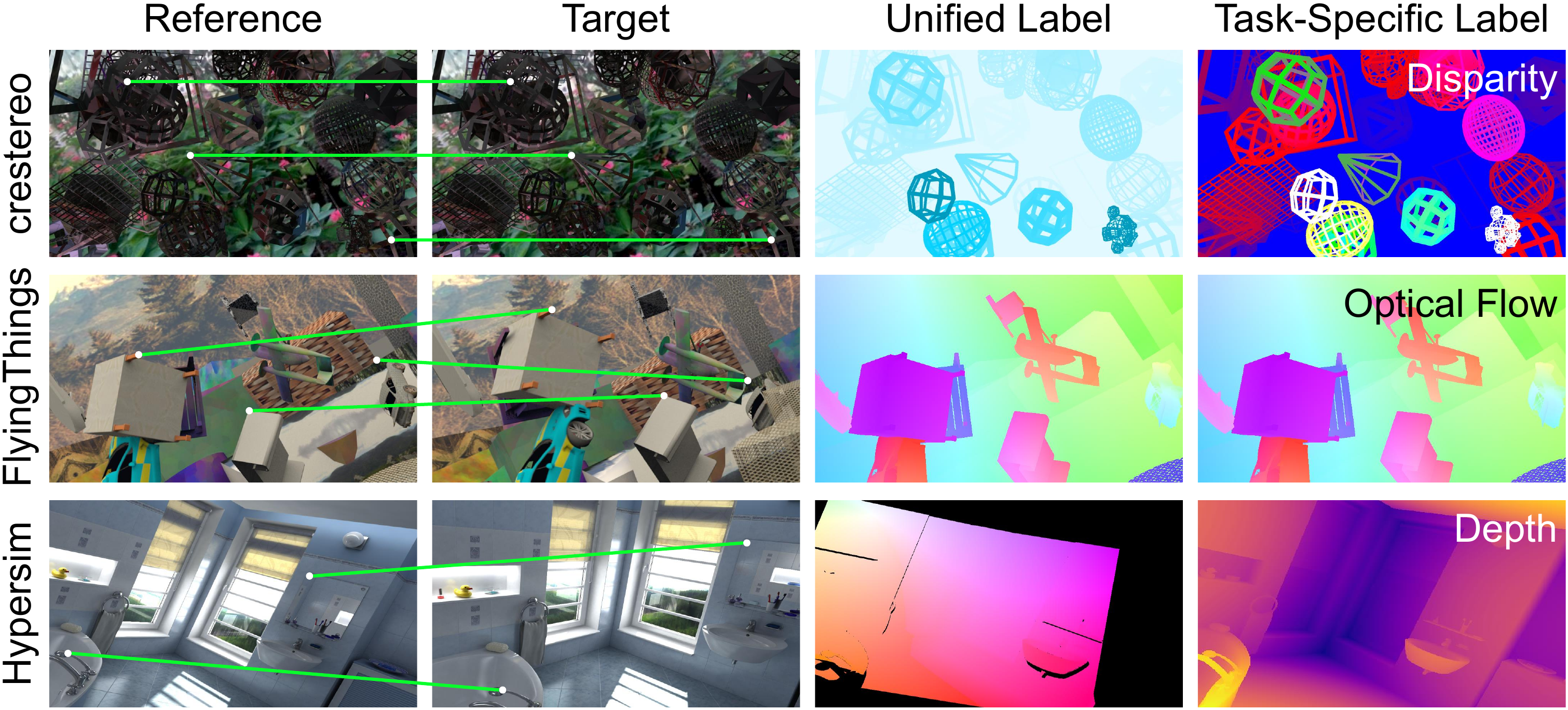}
    \caption{Diversity of the training data. Each row shows paired inputs in the first two columns with a few correspondences marked to illustrate their characteristics. 
    The corresponding ground-truth displacement in the third column is derived from the dataset’s original annotations in the fourth column.}
    \label{fig:adopted_datasets}
\end{figure}

% 我们的训练策略和数据集选择
% 数据集选择的写法参考 MONst3R
Unlike previous methods that are pretrained on small-scale datasets~\cite{FlowNet,DispNetC,Sintel}, we emphasize the importance of data diversity and collect a broad range of available optical flow datasets. Additionally, we incorporate datasets from feature matching and stereo matching tasks to further enrich the diversity of training samples.
% Since the supplementary samples lack optical flow annotations, we 
The annotations are obtained by converting ground-truth depth or disparity maps using camera parameters, as illustrated in Fig.~\ref{fig:adopted_datasets}. 
%we cannot obtain reliable, ground-truth optical flow on dynamic real-world scenes, even with human annotation~\cite{kubric}.
% Therefore, 
%Note that, we primarily use synthetic datasets for training, as the accurate optical flow, camera poses and depth information can be easily extracted during the rendering process.
The datasets adopted in our experiments are summarized in Table~\ref{tab:datasets}.
%Besides, we carefully selected the training data to ensure zero-shot evaluation on the benchmarks discussed in Sec.~\ref{sce:comparison_with_versatiles} \& \ref{sce:comparison_with_specifics}. 
% Note that, we exclude Driving~\cite{DispNetC}, IRS~\cite{irs}, and VIPER~\cite{viper} datasets that contain annotations inconsistent with real-world conditions (\textit{e.g.}, incorrect annotations on non-Lambertian surfaces).

\subsubsection{Training Strategies}
We use DINOv2-reg-giant~\cite{DINOv2, register} as the LVM encoder and select FlowFormer~\cite{FlowFormer} as the cross-view aggregation baseline to construct PanMatch. 
We employ a three-stage training strategy during the training phase.
\begin{itemize}
    \item Stage 1: We follow the training protocol in~\cite{FlowFormer} to establish a common optical flow baseline. %, pretraining for 100k iterations on Flying\textbf{C}hairs~\cite{FlowNet} and 120k iterations on Flying\textbf{T}hings3D~\cite{DispNetC}. 
    Images are randomly cropped to a size of $448 \times 448$. 
    \item Stage 2: The datasets listed in Table~\ref{tab:datasets} are included to train our PanMatch for 300k iterations with a batch size of 8. We keep the randomly cropped image to a size of $448 \times 448$ unchanged. 
    \item Stage 3: Only datasets shown in boldface in Table~\ref{tab:datasets} are included to fine-tune our PanMatch for 60k iterations. Specifically,  the cropped image size is increased to $672 \times 896$.
\end{itemize}
%we reduce the data sources to high-quality synthetic datasets, increasing the crop size to $672 \times 896$, and continue training for 60k iterations, aiming to maximize the model's ability to capture detailed structures.
% (3) In the third stage, we freeze the model and insert a learnable confidence head, training it for 50k iterations using high-quality synthetic datasets.
% 为什么要采用三阶段训练策略
{The core idea of our three-stage training strategy is to progressively learn displacement distributions, which starts from a single task, then moves to multiple tasks, and finally culminates in challenging fine-grained multi-task scenarios.}
% 学习率
Throughout all three stages, we use the AdamW~\cite{AdamW} optimizer and the one-cycle learning rate schedule, with the maximum learning rate being set to $1 \times 10^{-4}$.
% 数据增强
Data augmentations including color jitter, asymmetric occlusion and image flipping~\cite{RAFT} are conducted during training.
% 对 stereo 的数据后处理方法
For stereo pairs, %we additionally apply arbitrary vertical translations to the cropping window of the right view and randomly rotate the paired cropping window to interchange the $x$- and $y$-direction optical flow components. This approach mitigates bias introduced by zero optical flow and ensures balanced training of the optical flow components.
we randomly shift the right-view cropping window in the vertical direction, alleviating distribution imbalance caused by abundant zero displacement in stereo-based datasets. Besides, we randomly rotate the cropped stereo pair to interchange horizontal and vertical displacement components, promoting equitable training of both components.
% 对 posed pairs 的数据后处理方法
For posed images captured in rigid scenes~\cite{hypersim,megadepth}, we select adjacent poses with similar perspectives as posed pairs and convert their depth according to Eq.~\ref{eq:pinhole_camera_model} to obtain the corresponding displacement annotations for the same observation area, as illustrated in the $3rd$ row of Fig.~\ref{fig:adopted_datasets}.

\subsection{Comparison with Unified Models}
\label{sce:comparison_with_versatiles}
\subsubsection{Baselines}
% 我们是如何选择参与评测的方法的 (按照我们统一模型的构建，将光流方法直接构造成统一方法，并选择他们的in-the-wild checkpoint作为 evaluation 的结果，只考虑 stereo, depth, feature matching 的结果)
% We compare PanMatch with the optical flow model UniMatch~\cite{unimatch} and the feature matching models RoMa~\cite{RoMa}. 
% % All of these models demonstrate impressive performance in zero-shot downstream applications.
% UniMatch is selected due to its unified design and notable performance in the Robust Vision Challenge 2022.
% RoMa is chosen because of its capability to generate all-pairs correspondences suitable for optical flow and stereo matching evaluations, despite not being explicitly trained for these tasks.
% Both selected models support versatile outputs under multitask scenarios with minimal modifications.
% To ensure fairness, these two models are evaluated using their best checkpoints trained on available real-world datasets for in-the-wild scenarios. 
% Furthermore, a tiling strategy~\cite{PerceiverIO} is applied across all models  to optimize performance during stereo and optical flow evaluations.

With our unified formulation, existing optical flow estimation algorithms can be extended to unified matching models that support diverse outputs across multiple tasks. Consequently, we select four state-of-the-art optical flow algorithms for evaluation, including UniMatch~\cite{unimatch}, CrocoFlow~\cite{croco_v2}, SAMFlow~\cite{SAMFlow} and Flow-Anything~\cite{flow_anything}.
%UniMatch~\cite{unimatch} is selected due to its unified design and notable performance in the Robust Vision Challenge 2022. 
%Crocov2~\cite{croco_v2} is chosen because it was pretrained on large-scale correspondence datasets to learn generalizable matching priors. 
{Note that, both CrocoFlow and Flow-Anything employ million-scale datasets for self-supervised pretraining, while
SAMFlow~\cite{SAMFlow} leverages SAM~\cite{SAM} to enhance the accuracy of the estimated optical flow. }
%To ensure a fair comparison, all models are evaluated using their best checkpoints trained on real-world datasets for in-the-wild scenarios. 
%Additionally, a tiling strategy is applied across all models to optimize offset estimation accuracy for high-resolution inputs.

\subsubsection{Benchmarks and Metrics}
% 我们选择了哪些 benchmark, 绘制罗盘图并分析: unimatch, Crocov2
We select three tasks (including stereo matching, optical flow estimation and feature matching) to evaluate the performance of unified models in static scenes, dynamic scenes, and geometrically varying scenes, respectively.
% 立体匹配
For stereo matching, we assess disparity accuracy on the Middlebury (train-h)~\cite{Middlebury}, ETH3D (train)~\cite{ETH3D}, and KITTI 2012 (train)~\cite{KITTI2012} datasets. %, and DrivingStereo~\cite{DrivingStereo}, encompassing diverse scenarios including indoor environments, outdoor settings, and driving conditions under different weather variations. 
We use percentage of accuracy in threshold $x$ pixel (PCA $x$) as the evaluation metric, setting threshold 1px for ETH3D, 2px for Middlebury and 3px for KITTI, respectively.
% Evaluation metrics include the All-Bad $x$ Ratio for Middlebury and ETH3D, and the D1-All metric for KITTI 2012. % and the End-Point Error (EPE) metric for DrivingStereo datasets.
% 光流估计
For optical flow, benchmarks including Infinigen~\cite{infinigen} and Spring~\cite{spring} are employed, with PCA 1 being employed for evaluation.
For feature matching, we assess pose estimation accuracy on the ScanNet-1500~\cite{ScanNet} and YFCC100M~\cite{YFCC} datasets, and eigenvector estimation accuracy on the WxBS~\cite{WxBS} dataset.
Before feature matching evaluation, we employ a forward-backward circular consistency check to filter out correspondence outliers.
%Standard evaluation metrics are adopted for each dataset, and detailed definitions are omitted here for brevity. %It is worth noting that PanMatch demonstrates robust generalization across all benchmarks.

\subsubsection{Multi-Task Evaluation}
\begin{figure*}[t]
    \centering
    \includegraphics[width=1.0\textwidth]{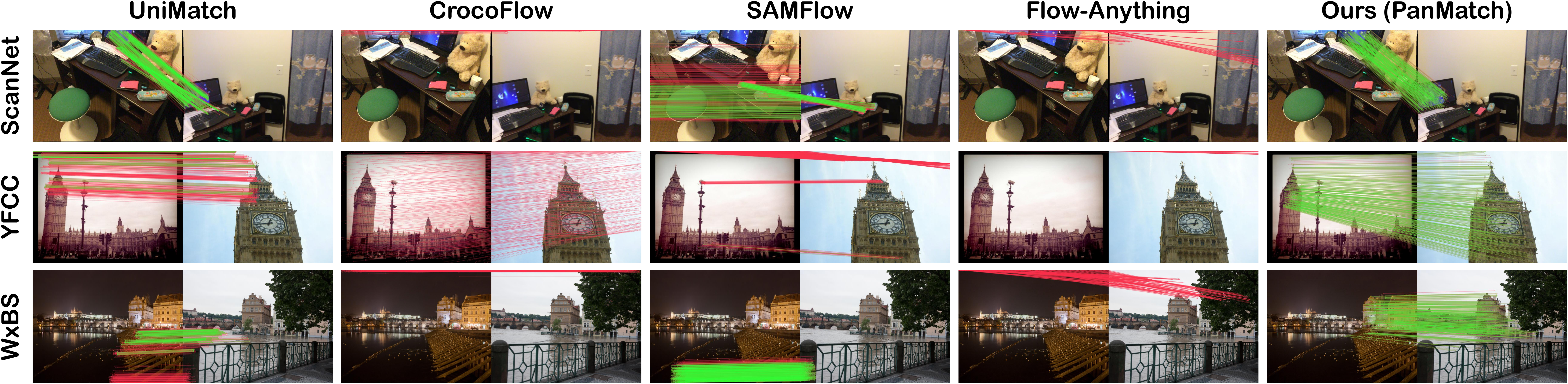}
    \caption{Qualitative results on the ScanNet~\cite{ScanNet}, YFCC~\cite{YFCC}, and WxBS~\cite{WxBS} datasets. All methods employ the same outlier filtering method and confidence threshold.}
    \label{fig:sparse_matching_comparison}
\end{figure*}

\begin{figure}[t]
    \centering
    \includegraphics[width=0.45\textwidth]{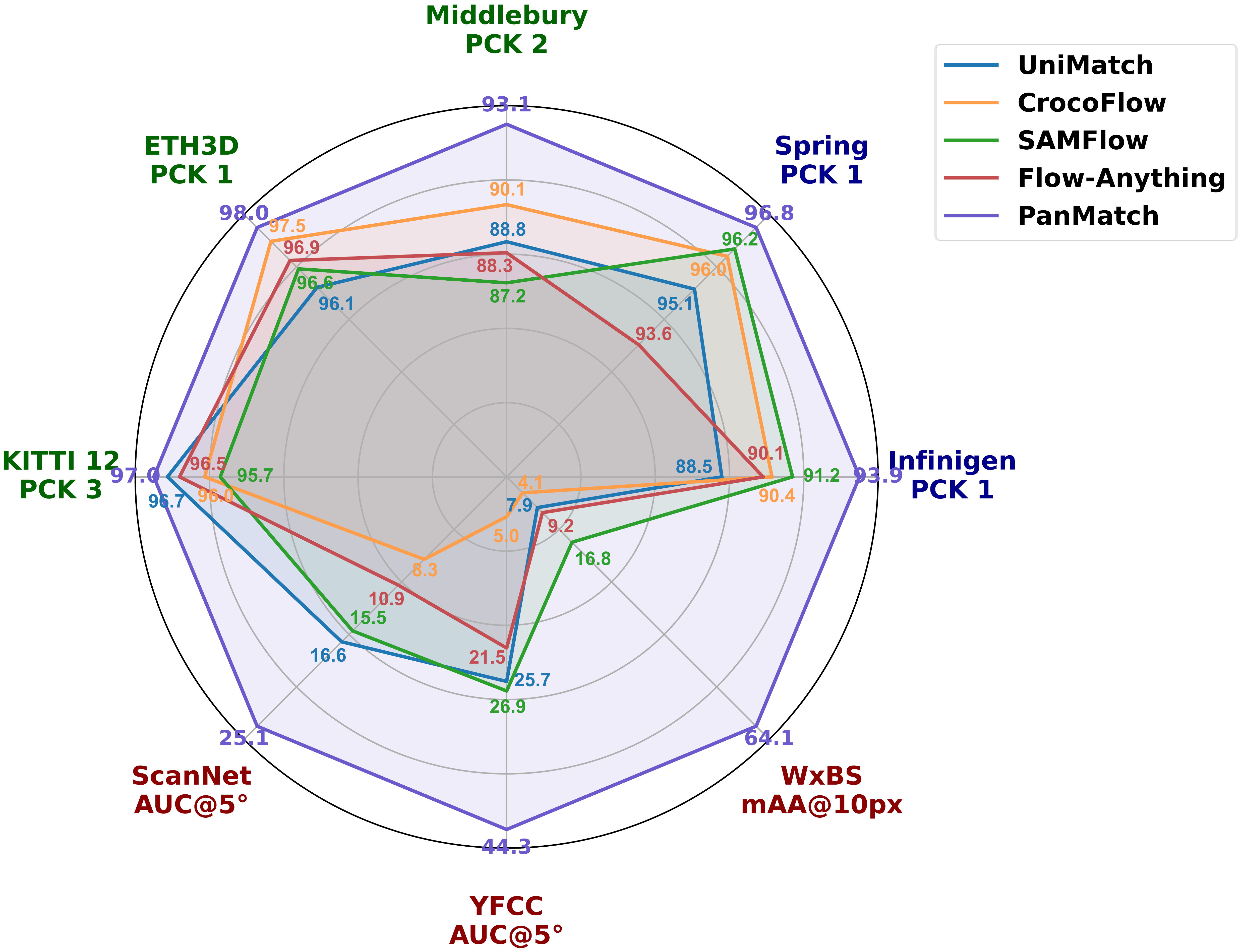}
    \caption{Zero-shot comparison on multiple domains and tasks, including {\color[rgb]{0,0,0.545098}optical flow estimation}, {\color[rgb]{0,0.39215686,0}stereo matching} and {\color[rgb]{0.545098,0,0}feature matching}.}
    \label{fig:multidomain_comparisons}
\end{figure}

\begin{figure}[t]
    \centering
    \includegraphics[width=0.45\textwidth]{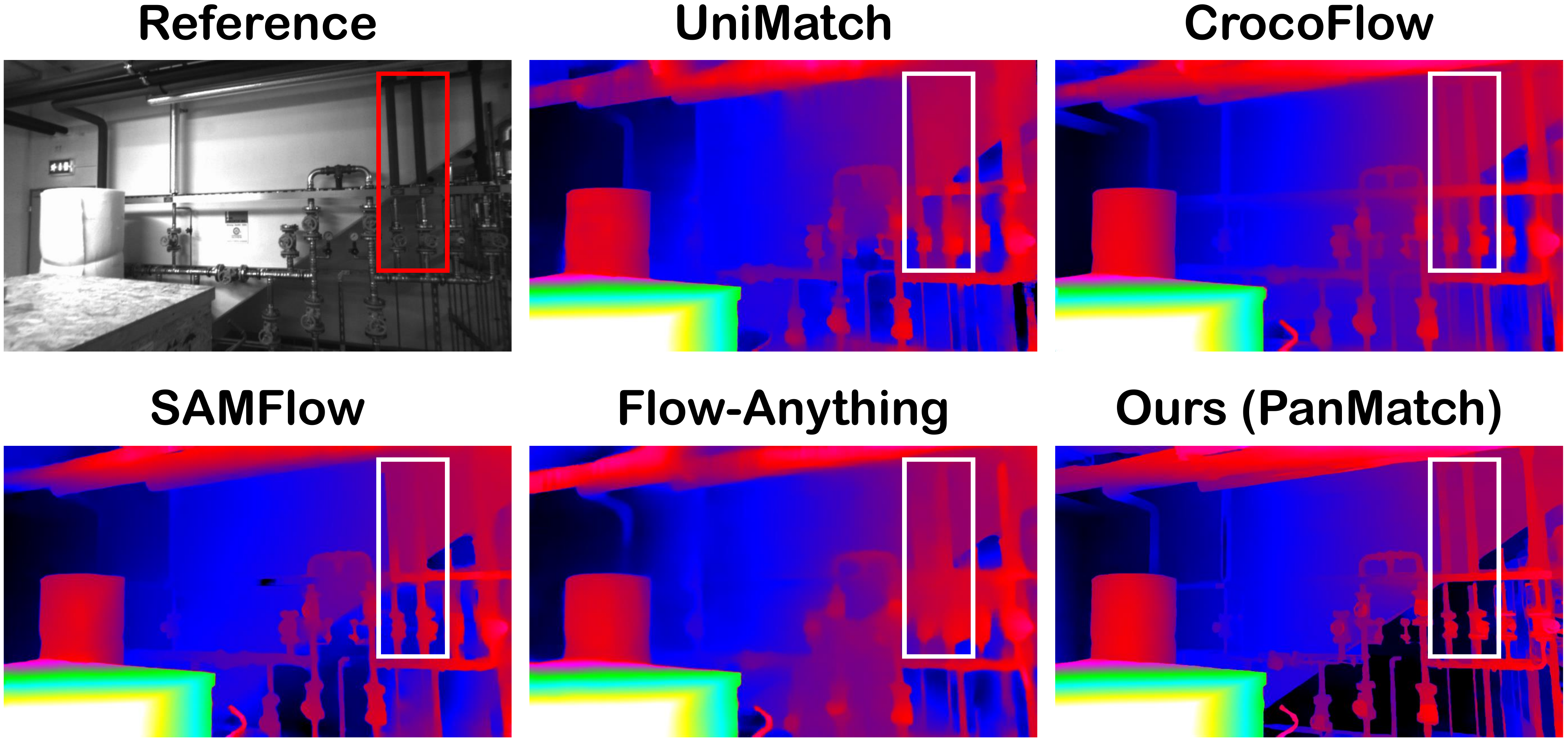}
    \caption{Comparison on an ETH test scene. Zoom out for better view.}
    \label{fig:ETH_comparison}
\end{figure}

\begin{figure*}[t]
    \centering
    \includegraphics[width=0.95\textwidth]{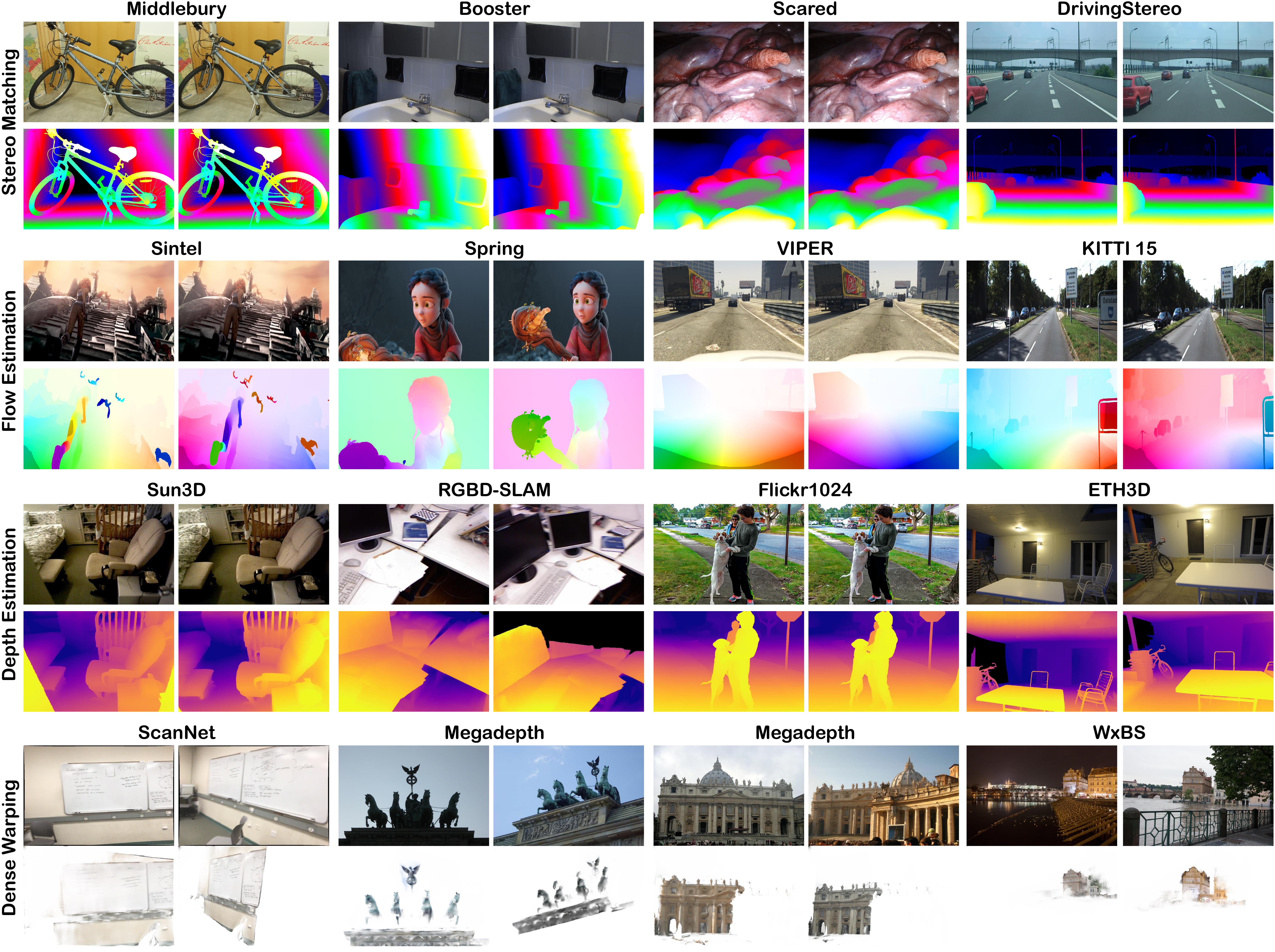}
    \caption{The versatile performance of PanMatch on a wide range of tasks and datasets~\cite{Middlebury,Booster,scared,DrivingStereo,Sintel,spring,viper,KITTI2015,sun3d,rgbd-slam,flickr1024,ETH3D,ScanNet,megadepth,WxBS}. Given a pair of two-view input, PanMatch can output disparity (group in the 1st row), optical flow (group in the 2nd row), relative depth (group in the 3rd row) and dense warping correspondences (group in the 4th row). Note that all test scenes are never seen during training.}
    \label{fig:vis_for_versatile}
\end{figure*}

\textbf{Optical Flow Estimation}.
As illustrated in Fig.~\ref{fig:multidomain_comparisons}, our approach achieves the highest accuracy on both the Infinigen and Spring validation dataset. Specifically, PanMatch attains 0.32 EPE on Infinigen and 0.31 EPE on Spring, leading 16\% and 23\% reductions compared to the second-best method (\textit{i.e.}, 0.38 EPE on Infinigen and 0.40 EPE on Spring).\\
\textbf{Stereo Matching}.
Our approach outperforms the second best competitor by 30\%, 16\%, and 10\% on the Middlebury, ETH3D, and KITTI12 datasets, respectively, underscoring its superior cross-task transferability. To facilitate an intuitive comparison, we present disparity estimates from an ETH3D test scene in Fig.~\ref{fig:ETH_comparison}. As illustrated, existing flow-based methods falter on complex, overlapping structures, yielding ambiguous foreground–background boundaries and blurred object contours, which impair disparity accuracy. In contrast, our method precisely delineates fine details and preserves subtle disparity variations.\\
\textbf{Feature Matching}.
{As illustrated in Fig.~\ref{fig:multidomain_comparisons}, existing flow-based methods struggle to perform feature matching tasks directly, exhibiting low performance compared to our approach. We provide a qualitative comparison in Fig.~\ref{fig:sparse_matching_comparison} for discussion. It can be observed that input pairs sampled from the ScanNet dataset exhibit large inter-frame motions and significant viewpoint changes, which limits the correspondence estimation accuracy for methods like CrocoFlow and Flow-Anything. Similarly, on the YFCC and WxBS datasets, drastic variations in appearance and pose yield few {usable} matches for all competing methods, severely impairing downstream tasks like pose and essential matrix estimation. In contrast, our approach remains robust under these challenging conditions, achieving significant performance gains against other methods. }\\
\textbf{Qualitative Evaluation}. 
We have visualized the zero-shot capability and multi-task versatility of our unified model on a broader range of tasks and datasets. As shown in Fig.~\ref{fig:vis_for_versatile}, our model can obtain reasonable outputs after a single feed-forward inference.

%--------------------------------------------------------------------------
\subsection{Comparison with Task-Specific Models}
\label{sce:comparison_with_specifics}
% 在各个 benchmarks 上的评测结果, 和各个专用方法的 RVC 版本对比. 但是RVC方法有限。于是构造stereo上的对比
\subsubsection{Stereo Matching and Optical Flow}
\textbf{Zero-shot Comparison}. 
% 在 stereo 数据集上对比 stereo
We evaluate zero-shot stereo matching performance on the half- and quarter-resolution subsets of Middlebury, the 2015 and 2012 training splits of KITTI, and the ETH3D training set.
Evaluation metrics include non-occluded Bad 2.0 for Middlebury, non-occluded Bad 1.0 for ETH3D, and D1-all for KITTI. 
We compare our method to state-of-the-art generalized stereo matching algorithms, as well as optical flow models fine-tuned on multiple real-world datasets.
{As shown in Table~\ref{tab:zero_shot_stereo_matching_evaluation}, RVC models~\cite{unimatch, MS-RAFT+} fine-tuned on real-world data significantly outperform those generalizable variants of PSMNet and RAFT-Stereo, indicating that pretraining on diverse data substantially improves cross-domain robustness. 
Furthermore, by pretraining on a large-scale, high-quality mixture of synthetic datasets, our PanMatch achieves the lowest zero-shot errors across all benchmarks, significantly outperforming FlowFormer++ and CrocoFlow that are pretrained on larger-scale but unlabeled data.
These comparisons underscore the importance of both the scale and quality of pretraining data.}
% Note that FlowFormer++ and CrocoFlow also leverage far larger but unlabeled datasets for pretraining. However, their performance is far inferior to ours, which demonstrates the value of high-quality annotated pairs.

% 在光流数据集上的 zero-shot 对比
For zero-shot optical flow estimation, we evaluate on the training subsets of Sintel and KITTI-Flow. 
Table~\ref{tab:zero_shot_optical_flow_evaluation} similarly shows that methods pretrained with additional data exhibit lower generalization errors, with our approach achieving the best results thanks to its use of large-scale annotated pairs.

% Please add the following required packages to your document preamble:
% \usepackage{multirow}
% \usepackage[normalem]{ulem}
% \useunder{\uline}{\ul}{}
\begin{table}[]
\setlength{\tabcolsep}{2pt}
\centering
\caption{Zero-shot comparison on stereo matching task. $\dag$ means testing on the training data.}
\scalebox{0.95}{
\begin{tabular}{llccccc}
\hline
                                                                                     &                                              & \multicolumn{2}{c}{Middlebury}                                                & \multicolumn{2}{c}{KITTI}                                                     &                                    \\ \cline{3-6}
\multirow{-2}{*}{Extra Data}                                                         & \multirow{-2}{*}{Method}                     & Half                                  & Quarter                               & 2015                                  & 2012                                  & \multirow{-2}{*}{ETH3D}            \\ \hline
                                                                                     & \cellcolor[HTML]{EFEFEF}PSMNet~\cite{PSMNet}               & \cellcolor[HTML]{EFEFEF}26.9          & \cellcolor[HTML]{EFEFEF}20.0          & \cellcolor[HTML]{EFEFEF}16.3          & \cellcolor[HTML]{EFEFEF}15.1          & \cellcolor[HTML]{EFEFEF}23.8       \\
                                                                                     & MS-PSMNet~\cite{MS-Net}                                    & 19.8                                  & 10.8                                  & 7.8                                   & 14.0                                  & 16.8                               \\
                                                                                     & FC-PSMNet~\cite{FCStereo}                                    & 15.1                                  & 9.3                                   & 5.8                                   & 5.3                                   & 9.5                                \\
                                                                                     & ITSA-PSMNet~\cite{ITSA}                                  & 12.7                                  & 9.6                                   & 5.8                                   & 5.2                                   & 9.8                                \\
                                                                                     & Graft-PSMNet~\cite{GraftNet}                                 & 9.7                                   & -                                     & 4.8                                   & 4.3                                   & 7.7                                \\
                                                                                     & HVT-PSMNet~\cite{HVT}                                   & 10.2                                  & -                                     & 4.9                                   & 4.3                                   & 6.9                                \\
                                                                                     & HODC-PSMNet~\cite{HODC-PSMNet}                                  & 9.3                                   & 7.0                                   & \underline{4.7}                             & 3.9                                   & 5.4                                \\
                                                                                     & ADL-PSMNet~\cite{ADLStereo}                                   & 8.9                                   & -                                     & 4.8                                   & 4.2                                   & \textbf{3.4}                       \\
                                                                                     & Former-PSMNet~\cite{FormerStereo}                                & \underline{7.7}                             & \underline{6.2}                             & 5.0                                   & \underline{3.9}                             & 6.7                                \\
                                                                                     & \cellcolor[HTML]{EFEFEF}\textbf{Ours+PSMNet} & \cellcolor[HTML]{EFEFEF}\textbf{7.63} & \cellcolor[HTML]{EFEFEF}\textbf{5.20} & \cellcolor[HTML]{EFEFEF}\textbf{4.68} & \cellcolor[HTML]{EFEFEF}\textbf{3.06} & \cellcolor[HTML]{EFEFEF}{\underline{5.09}} \\ \cline{2-7} 
                                                                                     & RAFT-Stereo~\cite{RAFT-Stereo}                                  & 12.6                                  & -                                     & 5.7                                   & 5.1                                   & 3.3                                \\
                                                                                     & HVT-RAFT~\cite{HVT}                                     & 10.4                                  & -                                     & 5.2                                   & 3.7                                   & 3.0                                \\
                                                                                     & Former-RAFT~\cite{FormerStereo}                                  & 8.1                                   & 5.4                                   & 5.1                                   & 3.9                                   & 3.3                                \\
                                                                                     % & \textbf{Ours+RAFT-S}                         & 8.68                                  & 5.61                                  & 5.59                                  & \textbf{3.74}                         & \textbf{2.21}                      \\ 
                                                                                     \hline
                                                                                     & GMFlow\_RVC~\cite{unimatch}                                  & 6.83                                  & 6.04                                  & 2.92$^\dag$                                  & 3.13$^\dag$                                  & 3.27                               \\
\multirow{-2}{*}{S+V+H+KITTI} & MS\_RAFT+\_RVC~\cite{MS-RAFT+}                               & 7.15                                  & 6.20                                  & 2.00$^\dag$                                  & 2.58$^\dag$                                  & 5.10                               \\
YouTube-VOS                                                                          & FlowFormer++~\cite{FlowFormer++}                                 & 10.99                                     & 10.21                                 & 6.12                                  & 4.50                                     & 4.63                               \\
CroCo-Pretrain                                                                       & CrocoFlow~\cite{croco_v2}                                    & 7.08                                  & -                                     & 4.19                                  & 3.83                                  & 2.59                               \\
Table~\ref{tab:datasets}                                                                              & \textbf{PanMatch}                        & \textbf{3.39}                         & \textbf{3.64}                         & \textbf{3.27}                         & \textbf{2.77}                         & \textbf{1.79}                      \\ \hline
\end{tabular}
}
\label{tab:zero_shot_stereo_matching_evaluation}
\end{table}

%--------------------------------------------------------------------------
% 我们的方法通用, 因此还设计了基于小数据集训练的方法对比
Note that our feature transformation pipeline can be seamlessly integrated into existing backbones, such as PSMNet, RAFT, and FlowFormer, by replacing their native feature extractors. 
We further validate its ability to enhance generalization when pretraining data are restricted.
As highlighted by the colored cells in Table~\ref{tab:zero_shot_stereo_matching_evaluation} and \ref{tab:zero_shot_optical_flow_evaluation}, integrating this module boosts generalization, yielding at least a 70\% generalization gain for PSMNet, 11\% for RAFT and 12\% for FlowFormer.
Notably, our integrated PSMNet outperforms other generalizable PSMNet variants under the same settings, underscoring the effectiveness of our architecture for robust generalization.

% Please add the following required packages to your document preamble:
% \usepackage{multirow}
% \usepackage[table,xcdraw]{xcolor}
% Beamer presentation requires \usepackage{colortbl} instead of \usepackage[table,xcdraw]{xcolor}
\begin{table}[]
\setlength{\tabcolsep}{3pt}
\centering
\caption{Zero-shot evaluation on optical flow estimation task. All models are pretrained with C+T scheduler.}
\scalebox{1.0}{
\begin{tabular}{clcccc}
\hline
                                             &                          & \multicolumn{2}{c}{Sintel}    & \multicolumn{2}{c}{KITTI}   \\ \cline{3-6} 
\multirow{-2}{*}{Extra Data}                 & \multirow{-2}{*}{Method} & Clean         & Final         & F1-epe     & F1-all         \\ \hline
                                             & PWC-Net~\cite{PWC-Net}                  & 2.55          & 3.93          & 10.4       & 33.7           \\
\rowcolor[HTML]{ECF4FF} 
                                             & RAFT~\cite{RAFT}                     & 1.43          & 2.71          & 5.04       & 17.4           \\
                                             & GMA~\cite{GMA}                      & 1.30          & 2.74          & 4.69       & 17.1           \\
                                             & SKFlow~\cite{SKFlow}                   & 1.22          & 2.46          & 4.27       & 15.5           \\
                                             & SEA-RAFT~\cite{SEA-RAFT}                 & 1.19          & 4.11          & 3.62       & 12.9           \\
                                             & GMFlow~\cite{GMFlow}                   & 1.08          & 2.48          & -          & -              \\
                                             & CCMR+~\cite{ccmr}                    & 0.98          & 2.36          & -          & 12.8           \\
\rowcolor[HTML]{EFEFEF} 
                                             & FlowFormer~\cite{FlowFormer}               & 0.94          & 2.33          & 4.09       & 14.7           \\
                                             & SAMFlow~\cite{SAMFlow}                  & 0.87          & 2.11          & 3.44       & 12.3           \\
\rowcolor[HTML]{ECF4FF} 
\multicolumn{1}{l}{\cellcolor[HTML]{ECF4FF}} & \textbf{Ours+RAFT}       & 1.27          & 2.29          & 3.76       & 12.58          \\
\rowcolor[HTML]{EFEFEF} 
                                             & \textbf{Ours+FlowFormer} & \textbf{0.83} & \textbf{2.05} & \textbf{3.10} & \textbf{12.12} \\ \hline
\multicolumn{1}{l}{MegaDepth}                & MatchFlow(G)~\cite{MatchFlow}             & 1.03          & 2.45          & 4.08       & 15.6           \\
\multicolumn{1}{l}{YouTube-VOS}              & FlowFormer++~\cite{FlowFormer++}             & 0.90          & 2.30          & 3.93       & 14.1           \\
\multicolumn{1}{l}{CroCo-Pretrain}           & CrocoFlow~\cite{croco_v2}                & -             & -             & 3.39       & 10.3           \\
\multicolumn{1}{l}{M+T+S}                    & RGM~\cite{RGM}                      & 0.90          & 2.40          & 3.90       & 12.5           \\
\multicolumn{1}{l}{Table~\ref{tab:datasets}} & \textbf{PanMatch}    & \textbf{0.85}    & \textbf{1.67} & \textbf{1.94} & \textbf{5.54}  \\ \hline
\end{tabular}
}
\label{tab:zero_shot_optical_flow_evaluation}
\end{table}

%-------------------------------------------------------------------------------
% Please add the following required packages to your document preamble:
% \usepackage{multirow}
\begin{table*}[]
    \setlength{\tabcolsep}{2pt}
    \caption{Robust vision challenges on stereo matching benchmarks.}
    \centering
    \scalebox{0.95}{
    \begin{tabular}{l|cccc|cccc|cccc}
    \hline
    \multirow{2}{*}{Method}          & \multicolumn{4}{c|}{Middlebury}                               & \multicolumn{4}{c|}{KITTI 2015}                                & \multicolumn{4}{c}{ETH3D} \\
                                     & \textbf{bad 2.0 (\%)} & \textbf{bad 4.0 (\%)} & AvgErr & Rank & \textbf{D1-bg (\%)} & \textbf{D1-fg (\%)} & D1-All (\%) & Rank & \textbf{bad 1.0 (\%)} & \textbf{bad 2.0 (\%)} & AvgErr & Rank \\
    \hline
    AANet\_RVC~\cite{AANet}                       & 31.8                  & 25.8                  & 12.8   & 13   & 2.23                & 4.89                & 2.67        & 11    & 5.41                  & 1.95                  & 0.33   & 11             \\
    GANet\_RVC~\cite{GANet}                       & 24.9                  & 16.3                  & 15.8   & 12   & 1.88                & 4.58                & 2.33        & 9    & 6.97                  & 1.25                  & 0.45   & 13            \\
    HSMNet\_RVC~\cite{HSMNet}                      & 16.5                  & 9.68                  & 3.44   & 6    & 2.74                & 8.73                & 3.74        & 13   & 4.40                  & 1.51                  & 0.28   & 10             \\
    MaskLacGwcNet\_RVC~\cite{MaskNet}               & 15.8                  & 10.3                  & 13.5   & 9    & 1.65                & 3.68                & 1.99        & 7    & 6.42                  & 1.88                  & 0.38   & 12            \\
    NLCANetV2\_RVC~\cite{NLCANetV2_RVC}                   & 16.4                  & 10.3                  & 5.60   & 8    & \underline{1.51}                & 3.97                & 1.92        & 5    & 4.11                  & 1.20                  & 0.29   & 9             \\
    Croco\_RVC~\cite{croco_v2}                       & 19.7                  & 12.2                  & 5.14   & 11    & 2.04                & 3.75                & 2.33        & 10    & 1.54                  & 0.50                  & 0.21   & 5             \\
    CFNet\_RVC~\cite{CFNet}                       & 16.1                  & 11.3                  & 5.07   & 7    & 1.65                & 3.53                & 1.96        & 6    & 3.70                  & 0.97                  & 0.26   & 8             \\
    UCFNet\_RVC~\cite{UCFNet}                      & 16.7                  & 10.9                  & 5.96   & 10    & 1.57                & 3.33                & 1.86        & 3    & 3.37                  & 0.78                  & 0.25   & 7             \\
    iRaftStereo\_RVC~\cite{iraftstereo_rvc}                 & 13.3                  & 8.02                  & 2.90   & 5    & 1.88                & 3.03                & 2.07        & 8    & 1.88                  & 0.55                  & 0.17   & 6             \\
    CREStereo++\_RVC~\cite{crestereo++}                 & 9.46                  & 6.25                  & 2.20   & 4    & 1.55                & 3.53                & 1.88        & 4    & 1.70                  & 0.37                  & 0.16   & 4             \\
    LoS\_RVC~\cite{LoS}                         & \underline{9.30}                  & 6.03                  & 2.36   & 3    & 1.58                & \underline{3.03}                & \underline{1.83}        & 2    & 1.47                  & \underline{0.25}                  & \underline{0.14}   & 3             \\
    DEFOM-Stereo\_RVC~\cite{DEFOM-Stereo}                & \textbf{6.90}                  & \textbf{4.25}                  & \textbf{1.61}   & 1    & \textbf{1.42}                & \textbf{2.68}                & \textbf{1.63}        & 1                              & \underline{1.09}               & 0.26   & \textbf{0.13}    & 2        \\
    % StereoAnything\_RVC              & -                     & -                     & -      &     & -                    & -                   & -           &      & 1.96                  & 1.07                 & 0.22    &   &     \\
    % FoundationStereo                 & 4.26                  & 2.72                  & 1.24   &     & -                    & -                   & -           &                       & 0.48               & 0.26   & 0.13    &   &     \\
    \textbf{PanMatch (Zero-shot)} & 11.4                 & \underline{5.71}                  & \underline{1.78}   & 2     & 2.74                & 6.26               & 3.33        & 12     & \textbf{0.61}                   & \textbf{0.14}                    & 0.15   & 1 \\
    \hline
    \end{tabular}
    }
    \label{tab:rvc_stereo}
\end{table*}

% Please add the following required packages to your document preamble:
% \usepackage{multirow}
\begin{table}[]
    \centering
    \caption{RVC comparison on optical flow benchmarks.}
    \scalebox{1.0}{
    \begin{tabular}{lcccc}
    \hline
    \multirow{2}{*}{Method} & KITTI         & Spring  & \multicolumn{2}{c}{Sintel}      \\ \cline{4-5} 
                            & F1-All        & Bad 1.0 & Clean          & Final          \\ \hline
    RAFT-TF\_RVC~\cite{RAFT}            & 5.56          & -       & 1.187          & 3.321          \\
    GMFlow\_RVC~\cite{unimatch}             & 4.41          & -       & \textbf{1.055} & \textbf{2.218} \\
    RAFT-it+\_RVC~\cite{RAFT-it+_RVC}           & \textbf{3.90} & -       & 1.837          & 2.696          \\
    MS\_RAFT+\_RVC~\cite{MS-RAFT+}          & \underline{4.15}          & \underline{5.724}   & 1.232          & 2.682          \\
    \textbf{PanMatch (Zero-shot)}   & 5.64          & \textbf{4.221}       & \underline{1.073}          & \underline{2.481}          \\ \hline
    \end{tabular}
    }
    \label{tab:rvc_flow}
\end{table}

% 和域泛化以及域自适应方法进行对比, 同时在与域泛化方法的对比过程中, 补充我们的其他基线在同样设置下的泛化性结果
\textbf{Robust Vision Challenges}.
Consistent with the objectives of this study, recent Robust Vision Challenges (RVC) aim at achieving robust performance across multiple benchmarks using a single model checkpoint. 
These approaches typically leverage multiple synthetic datasets for pretraining and various benchmark datasets for fine-tuning, thus significantly advancing real-world applications.
To evaluate the performance of PanMatch against these robust models, 
we submit its evaluation results to benchmarks and report the latest RVC leaderboard rankings for stereo matching and optical flow tasks in Table~\ref{tab:rvc_stereo} and Table~\ref{tab:rvc_flow}, respectively.
Note that PanMatch has neither been trained nor fine-tuned on any benchmark-related data, which is unlike competing methods that inherently benefit from such data to boost accuracy.
Nevertheless, PanMatch still outperforms most task-specific approaches.
Our PanMatch achieves the best performance on the ETH3D and Spring benchmark, while exhibits comparable performance on the Middlebury and Sintel benchmarks. 
% 解释原因
We hypothesize that the limited amount of 27 ETH3D training samples is insufficient for RVC methods to model in-domain characteristics. Consequently, model accuracy on this benchmark is dominated by zero-shot performance.
In contrast, KITTI, Middlebury and Sintel offer relatively abundant training data, allowing RVC methods to fit in-domain characteristics effectively and outperform our zero-shot model on these seen datasets.

% Please add the following required packages to your document preamble:
% \usepackage{multirow}
\begin{table}[]
    \setlength{\tabcolsep}{4pt}
    \centering
    \caption{Zero-shot comparison on Booster training set.}
    \scalebox{1.0}{
    \begin{tabular}{l|c|cccc}
    \hline
    \multirow{2}{*}{Method}                 & \multirow{2}{*}{Type}   & \multicolumn{4}{c}{Metric}                           \\ \cline{3-6} 
                                            &                         & \textbf{EPE}  & bad 2.0 & bad 3.0 & \textbf{bad 5.0} \\ \hline
    CFNet\_RVC~\cite{CFNet}                 & \multirow{4}{*}{Stereo} & 4.72          & 17.51   & 14.58   & 12.01            \\
    iRaftStereo\_RVC~\cite{RAFT-Stereo}     &                         & 4.01          & 13.21   & 11.17   & 9.47             \\
    LoS\_RVC~\cite{LoS}                     &                         & 2.36          & \textbf{10.74}   & 8.54    & 6.93             \\ \hline
    % DEFOM-Stereo\_RVC~\cite{DEFOM-Stereo}   &                         & 1.00          & 5.42    & 4.33    & 3.52             \\ \hline
    GMFlow\_RVC~\cite{unimatch}             & \multirow{3}{*}{Flow}   & 3.78          & 26.88   & 16.39   & 11.10            \\
    MS\_RAFT+\_RVC~\cite{MS-RAFT+}          &                         & 4.14          & 21.34   & 13.85   & 10.57            \\
    \textbf{PanMatch}            &                         & \textbf{1.95} & 17.45   & \textbf{6.82}    & \textbf{4.20}    \\ \hline
    \end{tabular}
    }
    \label{tab:rvc_booster}
\end{table}

\begin{figure*}[t]
    \centering
    \includegraphics[width=1.0\textwidth]{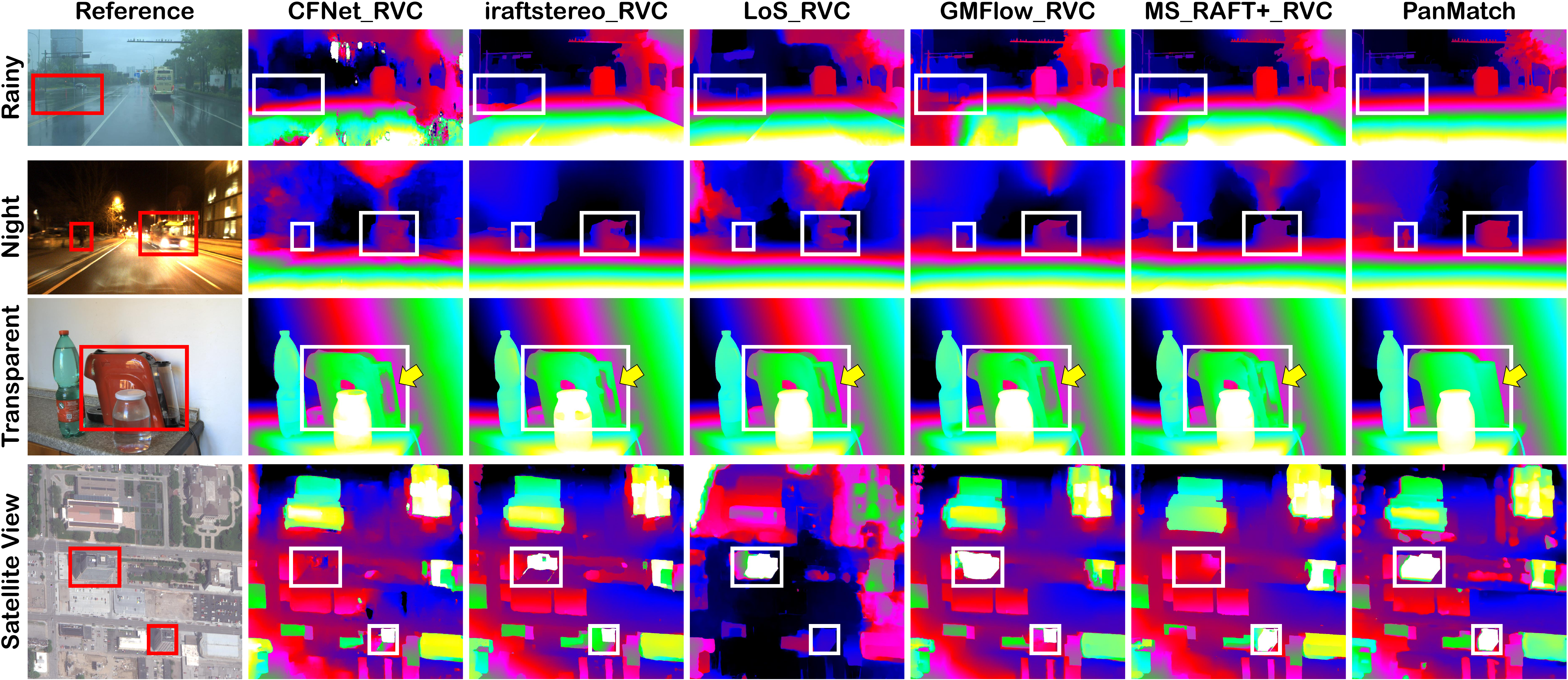}
    \caption{The real-world zero-shot capacity between RVC algorithms and our proposed PanMatch tested on four abnormal scenes, including rainy day of DrivingStereo~\cite{DrivingStereo}, night time of Oxford Robot Car~\cite{oxford}, transparent object in Booster~\cite{Booster} and satellite scene in UrbanSemantic3D~\cite{urbansemantic3d}. Challenging regions are marked by the white boxes. Zoom out for better view.}
    \label{fig:rvc_unseen_scenes}
\end{figure*}

% 通过数据说明引入额外数据对训练的增益, 比如将 hypersim + megadepth 移除, 测试在 booster 和 卫星图像的泛化结果
We argue that not ranking first on benchmark leaderboards does not necessarily indicate inferior real-world performance compared to task-specific methods fine-tuned with real-world datasets, as existing benchmarks do not fully represent real-world scenarios. 
Indeed, there exist challenging scenes where current RVC methods clearly lack robustness.
Specifically, we select the top 2 open-sources methods from the RVC stereo and optical flow leaderboard, \textit{i.e.}, iRaftStereo\_RVC~\cite{RAFT-Stereo,iraftstereo_rvc}, CFNet\_RVC~\cite{CFNet}, GMFlow\_RVC~\cite{unimatch}, MS-RAFT+\_RVC~\cite{MS-RAFT+}, as well as the recent LoS\_RVC~\cite{LoS}, % and DEFOM-Stereo\_RVC~\cite{DEFOM-Stereo}, 
and evaluate their performance on the Booster~\cite{Booster, Booster_PAMI} dataset.
% DrivingStereo captures outdoor driving scenes across four distinct weather conditions.
% Oxford Robot Car is an outdoor driving dataset captured in various time periods and seasons.
Booster is an indoor dataset characterized by diverse lighting conditions and challenging non-Lambertian materials such as mirrors and transparent windows.
We quantitatively evaluate all methods on the Booster dataset at quarter resolution (\textit{i.e.}, 1028$\times$752) in Table~\ref{tab:rvc_booster}, 
whose results indicating that these RVC methods struggle on previously unseen scenes and consequently lose their advantage over PanMatch.
Qualitative comparisons in Fig.~\ref{fig:rvc_unseen_scenes} further reveal that in challenging scenarios such as rainy day~\cite{DrivingStereo}, dark streets~\cite{oxford}, and satellite imagery~\cite{urbansemantic3d}, existing RVC algorithms fail to produce reliable results, whereas PanMatch consistently yields robust estimations.

% rvc方法在未见场景下的 zero-shot 能力

\subsubsection{Feature Matching}
% Please add the following required packages to your document preamble:
% \usepackage{multirow}
We evaluate PanMatch on widely adopted feature matching benchmarks, including ScanNet-1500~\cite{ScanNet}, MegaDepth-1500~\cite{megadepth}, and WxBS~\cite{WxBS}. 
% How to evaluate on the benchmarks.
Following the evaluation in RoMa~\cite{RoMa}, we assess pose estimation accuracy on ScanNet and MegaDepth, and compute the mean average accuracy (mAA) on ground truth correspondences consistent with the estimated fundamental matrix at a 10-pixel threshold for WxBS. 
To obtain reliable confidence estimation for PanMatch, we employ forward-backward cycle consistency in flow estimation to obtain the final confidence map. 
Results are presented in Table~\ref{tab:rvc_feature_matching}, which indicate that PanMatch achieves comparable performance on ScanNet even without fine-tuning on the specific indoor feature matching dataset. Besides, PanMatch outperforms most feature matching algorithms on the difficult WxBS benchmark and achieves the second best performance.
This is highly exciting since PanMatch is neither explicitly pretrained on the feature matching task nor designed to optimize correspondence selection for pose estimation~\cite{DKM,RoMa}.
% 增加一些措施来解决或者分析模型在Megadepth性能低下的原因
% Besides, we observe that PanMatch with joint confidence estimation strategy outperforms that relying solely on forward-backward cycle consistency.
% As illustrated in Fig.~\ref{fig:validate_confidence}, incorporating flow distribution uncertainty allows the model to more effectively filter out unreliable estimations compared to using forward-backward cycle consistency alone, thereby improving both pose and fundamental matrix estimation. 
% These observations highlight the effectiveness and importance of the proposed confidence estimation module.

\begin{table}[]
    \setlength{\tabcolsep}{2pt}
    \centering
    \caption{Comparisons on feature matching benchmarks. Higher metrics means better performance. $^\dagger$ means the model only uses forward-backward circle consistency as the confidence map.}
    \scalebox{1.0}{
    \begin{tabular}{l|ccc|ccc|c}
    \hline
    \multirow{2}{*}{Method}         & \multicolumn{3}{c|}{ScanNet-1500} & \multicolumn{3}{c|}{Megadepth-1500} & WxBS     \\
                                    & AUC@5      & 10        & 20       & AUC@5       & 10        & 20        & mAA@10 \\ \hline
    LoFTR~\cite{LoFTR}              & 22.1       & 40.8      & 57.6     & 52.8        & 69.2      & 81.2      & 55.4     \\
    PDC-Net+~\cite{PDC-Net+}        & 20.3       & 39.4      & 57.1     & 51.5        & 67.2      & 78.5      & -        \\
    ASpanFormer~\cite{ASpanFormer}  & 25.6       & 46.0      & 63.3     & 55.3        & 71.5      & 83.1      & -        \\
    DKM~\cite{DKM}                  & \underline{29.4}       & \underline{50.7}      & \underline{68.3}     & \underline{60.4}        & \underline{74.9}      & \underline{85.1}      & 58.9     \\
    RoMa~\cite{RoMa}                & \textbf{31.8}       & \textbf{53.4}      & \textbf{70.9}     & \textbf{62.6}        & \textbf{76.7}      & \textbf{86.3}      & \textbf{80.1}     \\
    \textbf{PanMatch}$^\dagger$          & 25.1       & 44.8      & 61.5     & 50.9        & 66.8      & 78.9      & \underline{64.2}     \\ \hline
    \end{tabular}
    }
    \label{tab:rvc_feature_matching}
\end{table}

\subsubsection{Depth Estimation}
\begin{table}[]
\setlength{\tabcolsep}{3pt}
\centering
\caption{Depth estimation results.}
\begin{tabular}{llcccc}
\hline
Dataset                    & Method                & Abs Rel        & Sq Rel         & RMSE           & RMSE log       \\ \hline
\multirow{6}{*}{RGBD-SLAM} & DeMoN~\cite{DeMoN}                 & 0.157          & 0.524          & 1.780          & 0.202          \\
                           & DeepMVS~\cite{DeepMVS}               & 0.294          & 0.430          & 0.868          & 0.351          \\
                           & DPSNet~\cite{DPSNet}                & 0.154          & 0.215          & 0.723          & 0.226          \\
                           & IIB~\cite{IIB}                   & \textbf{0.095} & -              & \textbf{0.550} & -              \\
                           & GMDepth~\cite{unimatch}               & 0.101          & 0.177          & 0.556          & 0.167          \\
                           & \textbf{PanMatch} & 0.123          & \textbf{0.134} & 0.568          & \textbf{0.190} \\ \hline
\multirow{6}{*}{SUN3D}     & DeMoN~\cite{DeMoN}                 & 0.214          & 1.120          & 2.421          & 0.206          \\
                           & DeepMVS~\cite{DeepMVS}               & 0.282          & 0.435          & 0.944          & 0.363          \\
                           & DPSNet~\cite{DPSNet}                & 0.147          & 0.107          & 0.427          & 0.191          \\
                           & IIB~\cite{IIB}                   & \textbf{0.099} & -              & \textbf{0.293} & -              \\
                           & GMDepth~\cite{unimatch}               & 0.122          & 0.068          & 0.336          & \textbf{0.146} \\
                           & \textbf{PanMatch} & 0.106          & \textbf{0.063} & 0.311          & 0.150          \\ \hline
\multirow{6}{*}{Scenes11}  & DeMoN~\cite{DeMoN}                 & 0.556          & 3.402          & 2.603          & 0.391          \\
                           & DeepMVS~\cite{DeepMVS}               & 0.210          & 0.373          & 0.891          & 0.270          \\
                           & DPSNet~\cite{DPSNet}                & 0.056          & 0.144          & 0.714          & 0.140          \\
                           & IIB~\cite{IIB}                   & 0.056          & -              & 0.523          & -              \\
                           & GMDepth~\cite{unimatch}               & 0.050          & 0.069          & 0.491          & 0.106          \\
                           & \textbf{PanMatch} & \textbf{0.031} & \textbf{0.029} & \textbf{0.307} & \textbf{0.083} \\ \hline
\multirow{3}{*}{ScanNet}   & DeMoN~\cite{DeMoN}                 & 0.231          & 0.520          & 0.761          & 0.289          \\
                           & GMDepth~\cite{unimatch}               & \textbf{0.059} & 0.019          & 0.169          & \textbf{0.080} \\
                           & \textbf{PanMatch} & 0.064          & \textbf{0.019} & \textbf{0.148} & 0.090          \\ \hline
\end{tabular}
\label{tab:rvc_depth}
\end{table}

As demonstrated in Sec.~\ref{sec:conversion_mechanism}, our unified model can output metric depth from the estimated flow once the camera parameters are available. 
% These requirements are identical to those of the unrectified two-view depth estimation task.
To evaluate the depth estimation accuracy of PanMatch, we test it on four depth benchmarks, including RGBD-SLAM~\cite{rgbd-slam}, Sun3D~\cite{sun3d}, Scenes11~\cite{DeMoN} and ScanNet~\cite{ScanNet}.
Following unrectified two-view depth estimation methods, we use four error metrics to assess depth quality, including absolute relative difference (Abs Rel), squared relative difference (Sq Rel), root mean squared error (RMSE) and RMSE in log scale (RMSE log). 
As shown in Table~\ref{tab:rvc_depth}, PanMatch achieves comparable performance across most metrics, and outperforms existing methods in terms of Sq Rel across four datasets, despite not being fine-tuned for the depth regression task. 
These results highlight the effectiveness and practicality of displacement-based depth estimation, and demonstrate the strong cross-task generalization capacity of PanMatch.

\begin{figure}[t]
    \centering
    \includegraphics[width=0.45\textwidth]{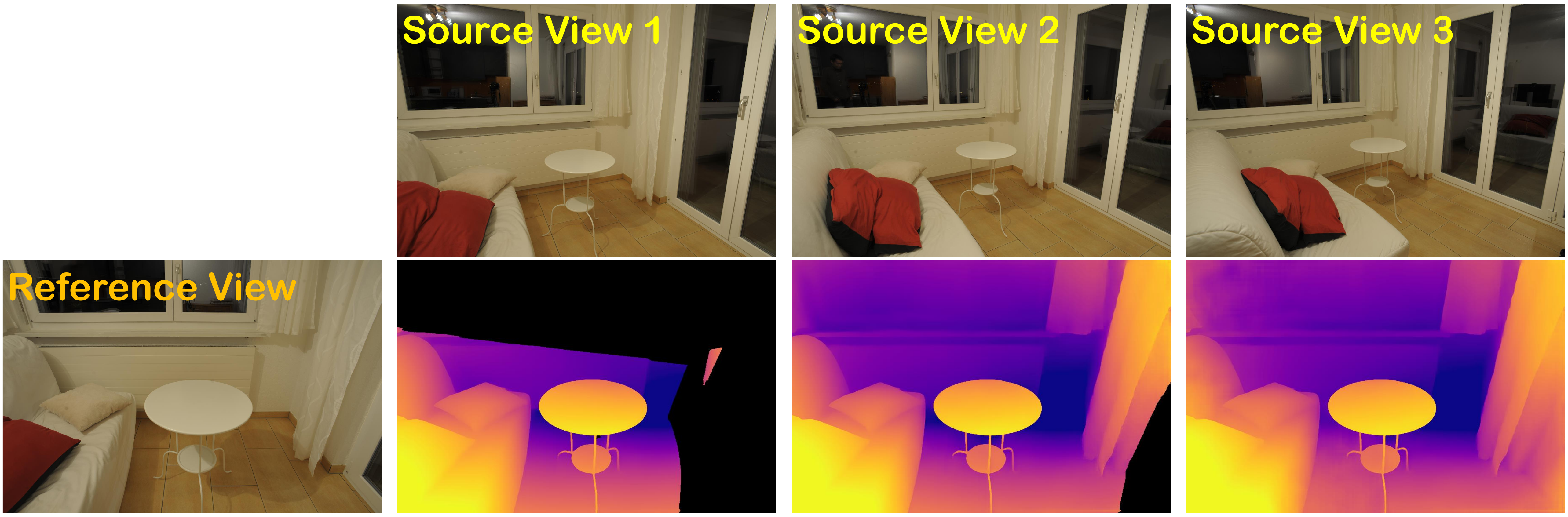}
    \caption{Two-view depth estimation. Each result comes from the same reference image and adjacent source inputs. Slightly changing the camera pose can find a suitable viewpoint to obtain a complete depth estimation map.}
    \label{fig:depth_by_views}
\end{figure}

Note that we only assess the valid regions of each estimated depth map
since our model may output a semi-dense depth map after depth validity check, which is adopted to filter potential numerical issues caused by division by zero in Eq.~\ref{eq:flow2depth_method1} or Eq.~\ref{eq:flow2depth_method2}.
Such situations occur when camera motion exhibits zero translation or is purely along the optical axis.
% We give the average output density of each datasets as follow: 34.3\% for RGBD-SLAM, 29.6\% for Sun3D, 56.4\% for scenes11 and 48.5\% for ScanNet. 
This does not point a weakness, as the semi-density issues can be easily alleviated by slightly adjusting the camera pose, which is practically achievable as illustrated in Fig.~\ref{fig:depth_by_views}. 
Additionally, we highlight the strong zero-shot generalization advantage of PanMatch, which makes it more suitable for real-world applications.
Furthermore, since PanMatch can estimate the relative pose from reliable correspondeces, it reduces the need for prior pose information, showing strong convenience for depth estimation in real-world scenarios compared to unrectified two-view algorithms.

\subsection{Model Analyses}
\label{sec:model_analyses}
\subsubsection{Ablation Study}
\label{sec:model_ablation}
We conduct experiments to validate the contributions of each block designed in the feature transformation pipeline. For simplicity, all ablation study defaultly use DAMv2-Base as the LVM encoder, and FlowFormer as the cross-matching baseline. We using C+T training schedular~\cite{PWC-Net} and approximately assess the real-world generalization ability by evaluating these variants on the training sets of the Sintel, KITTI and Middlebury datasets. 

% 以 stereo 任务作为 ablation baseline

\begin{figure}[t]
    \centering
    \includegraphics[width=0.45\textwidth]{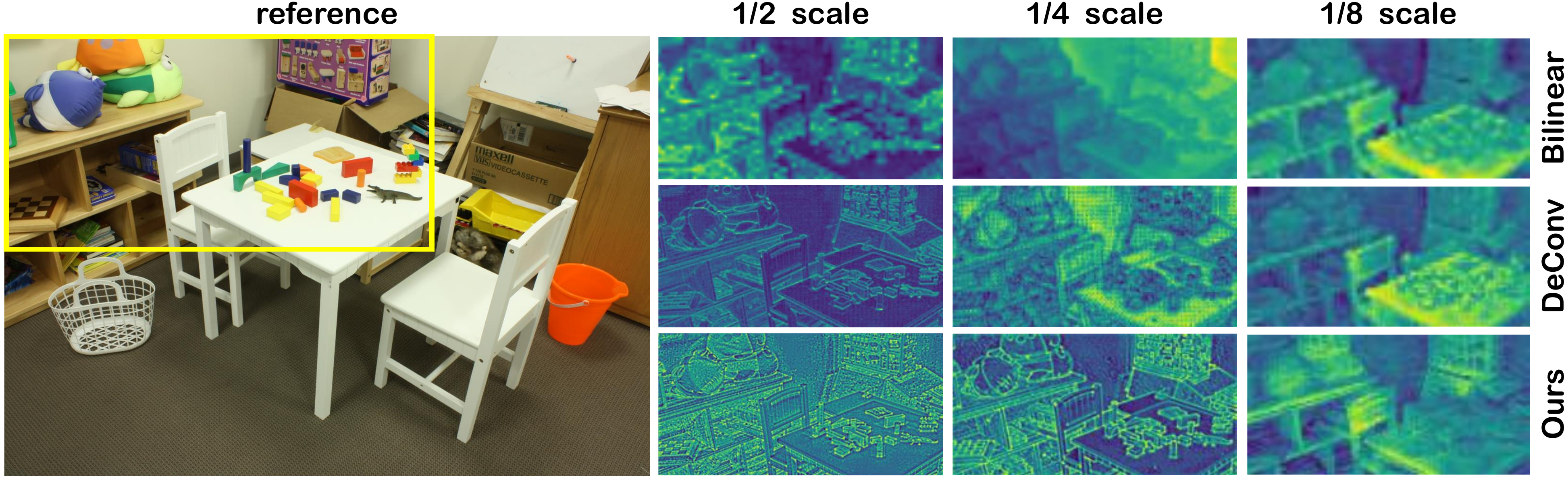}
    \caption{PCA analysis.}
    \label{fig:PCA_analysis}
\end{figure}

\textbf{Guided Feature Upsampling.}
% 为什么要提出这个新的上采样方案，相比双线性、deconvolution和pixel shuffle有什么优势 (这个问题似乎在 method 部分已经回答了, 即我们提出 guided upsampling block 的动机. 所以感觉不需要在这里再次重复)
% Patch embedding in ViT compresses the image resolution into 1/14, causing the loss of local details.
% As a result, an appropriate rescaling strategy is necessary to reconstruct the lost information for accurate dense matching.
In PanMatch, we propose a guided feature upsampling block to recompose feature maps at arbitrary scales under appropriate guidance. This module can be replaced by standard interpolation methods, such as bilinear interpolation, deconvolution, or pixel shuffling, once the target scale is known.
We constructed network variants employing these interpolation methods, investigating the necessity and effectiveness of the proposed block.
% 下面的分析应该从泛化性的角度去考虑, 从而与我们的动机相呼应
{As shown in Table~\ref{tab:ablation_guided_upsampling}, our proposed block achieves the best generalization performance on the Sintel, KITTI and Middlebury datasets.
To investigate what causes the differences among these upsampling strategies in terms of domain generalization, we present Principal Component Analysis (PCA) visualizations along the channel dimension of the upsampled features for intuitive comparison.
As illustrated in Fig.~\ref{fig:PCA_analysis}, bilinear interpolation fails to recover detailed information such as edges and small objects in the feature maps, resulting in increasingly blurry features as the required scale grows.
Although deconvolution generates sharp contours in high-resolution features, it also introduces checkerboard artifacts, particularly in textureless regions. These artifacts can lead to multi-modal matching distributions and hinder accurate target regression.
In contrast, our proposed guided feature upsampling method restores scene details and semantics with clear representations across scales, demonstrating strong generalization capabilities and ultimately enabling precise 2D displacement estimation.}
% These results demonstrate the superior effectiveness of the proposed guided upsampling module, facilitating enhanced generalization and improved matching performance.

% Please add the following required packages to your document preamble:
% \usepackage{multirow}
\begin{table}[]
    \centering
    \caption{Ablation study on interpolation strategy.}
    \begin{tabular}{lcccc}
    \hline
    Interpolation
                                   & \begin{tabular}[c]{@{}c@{}}Sintel\\ clean\end{tabular} & \begin{tabular}[c]{@{}c@{}}Sintel\\ final\end{tabular} & \begin{tabular}[c]{@{}c@{}}KITTI\\ EPE\end{tabular} & \begin{tabular}[c]{@{}c@{}}Middlebury\\ Bad 2.0\end{tabular} \\ \hline
    Bilinear                       & 1.05                                                & 2.26                                               & 4.48                                                & 14.89                                                 \\
    Deconvolution                  & 1.02                                                & 2.10                                                & 3.83                                                & 12.83                                                 \\
    Pixel shuffling                & 0.95                                                & 2.12                                                & 3.39                                                & 11.29                                                 \\
    \textbf{Ours}                  & \textbf{0.89}                                                & \textbf{2.07}                                                & \textbf{3.37}                                                & \textbf{10.86}                                                  \\ \hline
    \end{tabular}
    \label{tab:ablation_guided_upsampling}
\end{table}

\begin{figure}[t]
    \centering
    \includegraphics[width=0.45\textwidth]{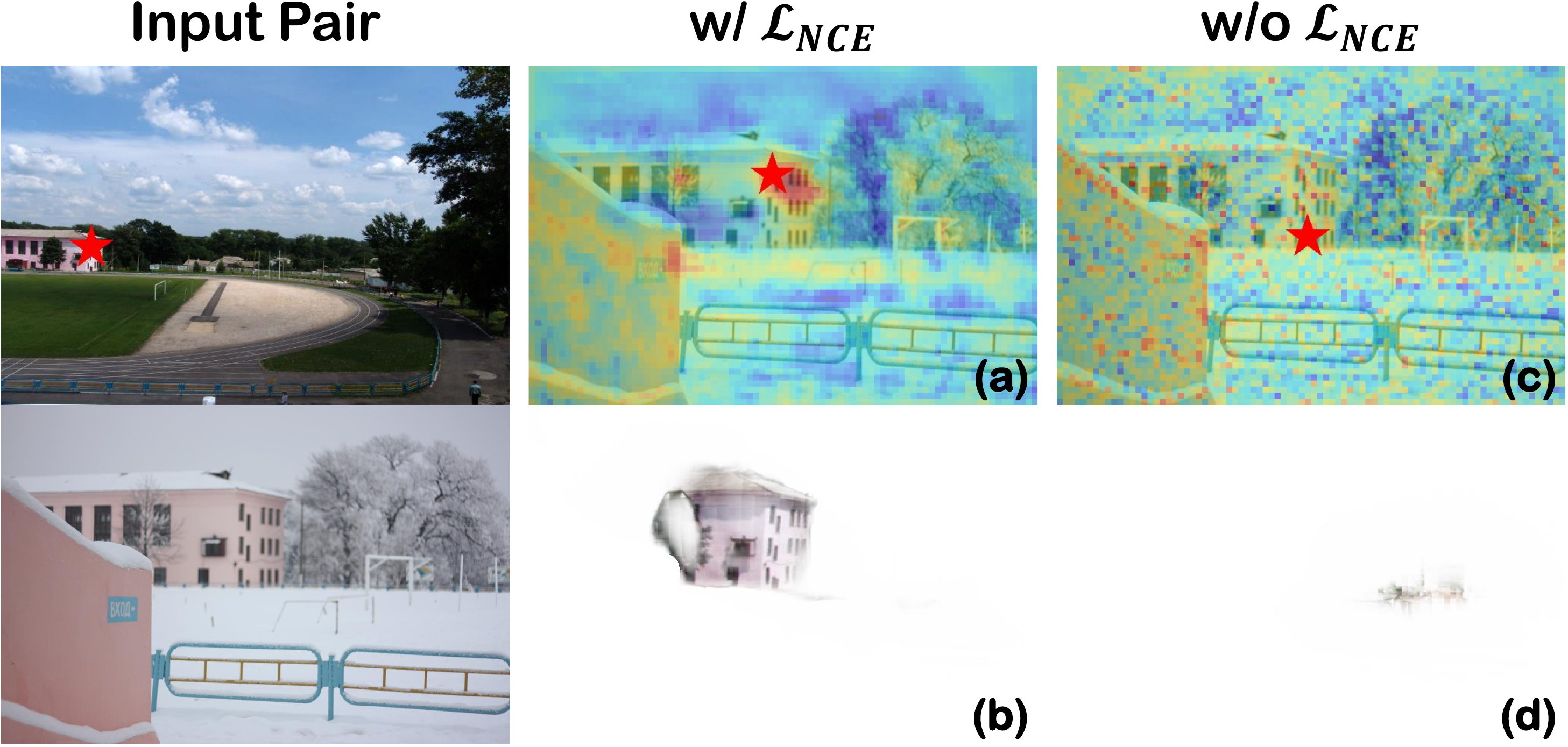}
    \caption{Generalization enhancement by $\mathcal{L}_{NCE}$. 
    % Given a pair of image with significant viewpoint and appearance difference, as well as the query point marked by red star, model trained with $\mathcal{L}_{NCE}$ can output a focus response on the correspondence position, and finally regress the correct position. The model without $\mathcal{L}_{NCE}$, output a noise response map and fail to regress the correspondence position. 
    Here, Figs. (a) and (c) are heatmaps generated by feature matching, and (b) and (d) are warped images from another view.}
    \label{fig:ablation_infoNCE}
\end{figure}

\textbf{Network Design and Auxiliary Losses.}
% 多层级融合是否是必须的？直接使用 Guide feature upsampling 得到 1/4 或者 1/8 的特征不行吗？
% 为什么需要多尺度信息融合？除了不浪费上一阶段辛辛苦苦提取的多尺度特征外，还能更进一步地更直接地利用多尺度信息，避免多尺度信息在级联结构中被遗忘（宽网络和深度网络的组合拳）
% 加入多尺度重建和语义一致性的好处
The proposed adapter employs a shallow Feature Pyramid Network (FPN) to facilitate task-specific adaptation and multi-scale feature fusion. Subsequently, Multi-scale Patch Embedding (MPE) integrates these multi-scale features into unified representations, which are ultimately supervised using an InfoNCE loss $\mathcal{L}_{NCE}$.
To systematically evaluate the contributions of each component toward model generalization, we first establish a baseline model composed of the LVM encoder, a guided upsampling module, and a cross-matching algorithm. Then, we incrementally introduce the FPN, MPE, and $\mathcal{L}_{NCE}$. As demonstrated in Table~\ref{tab:module_ablation}, progressively incorporating these components significantly enhances the model's generalization performance across multiple datasets. Specifically, introducing FPN and MPE enables the model to effectively utilize information from multiple layers of the ViT encoder, substantially improving both estimation accuracy and generalization capability. These findings align with the conclusions drawn from the toy experiments in Sec.~\ref{sec:toy_example}. Additionally, the integration of the InfoNCE loss enhances the discriminative capability of the representations utilized to construct the cost volume, thereby improving the model's generalization.
This hypothesis is validated in Fig.~\ref{fig:ablation_infoNCE}, which visualizes feature-matching results. Given a query point (indicated by a red star), the model trained with the InfoNCE loss exhibits cost-volume responses concentrated accurately around the correct corresponding location in the target view. Consequently, the predicted optical flow is highly accurate, enabling effective dense warping within the co-visible region, as illustrated by the clearly warped buildings in Fig.~\ref{fig:ablation_infoNCE} (b). In contrast, the model trained without the InfoNCE loss yields dispersed responses throughout the image, leading to an incorrect regression position for the query point and ineffective dense warps, %and unclear, unreliable warped 
as shown in Fig.~\ref{fig:ablation_infoNCE} (c) and (d).

% \begin{table}[]
%     \centering
%     \caption{We ablate network design and loss components following the C+T generalization evaluation.}
%     \begin{tabular}{|cc|cc|cccc|}
%     \hline
%     \multicolumn{2}{|c|}{Network} & \multicolumn{2}{c|}{Loss} & \multicolumn{4}{c|}{Evaluation} \\ \hline
%     FPN           & MPE           & $\mathcal{L}_{contex}$  & $\mathcal{L}_{recon}$ & S-c   & S-f   & K     & M     \\ \hline
%     \XSolidBrush  & \checkmark    & \XSolidBrush            & 0                     & 1.43  & 2.40  & 14.59 & 16.78 \\
%     \checkmark    & \XSolidBrush  & \XSolidBrush            & 0                     & 1.46  & 2.98  & 16.88 & 19.83 \\ \hline
%     \checkmark    & \checkmark    & \XSolidBrush            & 4                     & 0.91  & 2.04  & 12.84 & 11.47  \\ 
%     \checkmark    & \checkmark    & \checkmark              & 1                     &       &       &       &       \\
%     \checkmark    & \checkmark    & \checkmark              & 0                     & 0.89  & 2.16  & 12.57 & 12.25 \\ \hline
%     \checkmark    & \checkmark    & \checkmark              & 4                     &       &       &       &       \\ \hline
%     \ding{52}    & \ding{52}    & \ding{56}              & 0                     & 0.89  & 2.07  & 12.41 & 10.86  \\ \hline
%     \end{tabular}
%     \label{}
% \end{table}

\begin{table}[]
    \setlength{\tabcolsep}{6pt}
    \centering
    \caption{Ablation study on module design. Lower is better.}
    \scalebox{1.0}{
    \begin{tabular}{ccccccc}
    \hline
    FPN        & MEP        & $\mathcal{L}_{NCE}$   & \begin{tabular}[c]{@{}c@{}}Sintel\\ clean\end{tabular} & \begin{tabular}[c]{@{}c@{}}Sintel\\ final\end{tabular} & \begin{tabular}[c]{@{}c@{}}KITTI\\ F1-All\end{tabular} & \begin{tabular}[c]{@{}c@{}}Middlebury\\ Bad 2.0\end{tabular} \\ \hline
    \textcolor{red}{\ding{56}}  & \textcolor{red}{\ding{56}}  & \textcolor{red}{\ding{56}}             & 1.08                                                & 2.11                                                & 13.93                                                & 15.09                                                 \\
    \textcolor{green}{\ding{52}}  & \textcolor{red}{\ding{56}}  & \textcolor{red}{\ding{56}}             & 1.04                                                & 2.05                                                & 13.73                                               & 14.86                                                 \\
    \textcolor{green}{\ding{52}}  & \textcolor{green}{\ding{52}}  & \textcolor{red}{\ding{56}}             & 0.89                                                & 2.07                                                & 12.41                                                & 10.86                                                 \\
    \textcolor{green}{\ding{52}}  & \textcolor{green}{\ding{52}}  & \textcolor{green}{\ding{52}}             & \textbf{0.83}                                                & \textbf{2.05}                                                & \textbf{12.12}                                                & \textbf{10.04}                                                 \\ \hline
    \end{tabular}
    }
    \label{tab:module_ablation}
\end{table}

\subsubsection{Contributions from Architectures and Datasets}
\label{sec:data_ablation}
We attribute the powerful zero-shot generalization capability of PanMatch to two key factors: (1) domain-robust LVM representations, and (2) cross-task multi-domain training data. To validate this hypothesis, we systematically examine how these factors individually contribute to the baseline method's performance on multiple benchmark matching tasks. 
% For simplicity, our evaluation focuses on optical flow generalization and cross-task adaptability in stereo matching, excluding feature matching since the baseline lacks confidence estimation. 
% Our evaluation focuses on optical flow generalization and cross-task adaptability in stereo matching and feature matching. 
In current study, DINOv2-reg-giant is adopted as the LVM encoder, and FlowFormer is selected as the unified matching baseline.
%Note that we only employ forward-backward circle consistency for feature matching, avoiding the inconsistent impact from confidence estimation.

We conduct experiments to investigate: (i) whether LVM representations enhance domain generalization and cross-task adaptability when being trained on the same-scale dataset, (ii) whether the inclusion of additional optical flow data improves the baseline’s generalization, and (iii) whether incorporating cross-task data continues to benefit performance even when sufficient optical flow data (\textit{i.e.}, 1M samples) is available.  

\begin{table}[]
    \setlength{\tabcolsep}{2pt}
    \centering
    \caption{Comparison with the baseline under different training samples and feature backbone settings. Lower is better except for $\mathrm{mAA}$. \textit{OF} represents the collected flow datasets in Table~\ref{tab:datasets}.}
    \scalebox{1.0}{
    \begin{tabular}{ccc|cc|cc|c}
    \hline
    \multirow{3}{*}{ID} & \multirow{3}{*}{Datasets} & \multirow{3}{*}{Extractor} & \multicolumn{2}{c|}{Flow} & \multicolumn{2}{c|}{Stereo} & Match  \\
                        &                           &                            & Sintel        & KITTI         & Booster       & Middlebury    & WxBS   \\
                        &                           &                            & EPE           & F1-All        & EPE           & Bad 2.0       & mAA@10 \\ \hline
    1                   & C+T                       & SVT-L                      & 2.40          & 14.72         & 5.28          & 15.91         & 10.1   \\
    2                   & C+T+OF                    & SVT-L                      & 2.29          & 8.28          & 5.10          & 14.51         & 12.3   \\
    3                   & Table~\ref{tab:datasets}  & SVT-L                      & 2.55          & 8.09          & 3.48          & 14.08         & 13.3   \\ \hline
    4                   & C+T                       & Ours                       & 1.94          & 13.39         & 4.13          & 12.25         & 29.4   \\
    5                   & C+T+OF                    & Ours                       & 1.69          & 5.68          & 3.18          & 8.94         & 16.5       \\
    6                   & Table~\ref{tab:datasets}  & Ours                       & \textbf{1.67} & \textbf{5.54} & \textbf{1.95} & \textbf{6.88} & \textbf{64.2}   \\ \hline
    \end{tabular}
    }
    \label{tab:variants}
\end{table}

We answer these questions through quantitative evaluations on the optical flow datasets~\cite{Sintel,KITTI2015}, stereo matching datasets~\cite{Booster,Middlebury} and feature matching benchmark~\cite{WxBS}. 
{Table~\ref{tab:variants} compares the baseline method with its LVM-augmented variant (Model 1 \textit{vs.} 4, 2 \textit{vs}. 5, and 3 \textit{vs}. 6) on the same-scale datasets. These results demonstrate that LVM representations consistently enhance performance, not only for the original optical flow task but also in cross-task generalization.
% The advantages of employing LVM representations are especially notable when the training data are limited. 
Besides, this advantage maintains across all training data scales.
Moreover, while Model 3 exhibits degradation on Sintel and minimal improvement on WxBS, its LVM-augmented counterpart, Model 6, achieves significant gains across all evaluations. This contrast suggests that the baseline's encoder struggles to learn general features from multi-task data, limiting overall performance. In contrast, LVM provides domain-invariant features, enabling broader data to further boost cross-task generalization for the matching module of baseline method.}
% Moreover, even when the baseline is trained on large-scale datasets and already exhibits strong cross-task performance (model 3), incorporating LVM representations (model 6) still consistently outperforms the baseline, highlighting the effectiveness of LVM representations under both limited and abundant data conditions. 

{On the other hand, comparisons between Models 1 \& 2 or 4 \& 5 indicate that incorporating larger amounts of optical flow data notably enhances optical flow and stereo matching accuracy. However, feature matching accuracy remains suboptimal, despite employing sufficient optical flow data (\textit{e.g.}, 1M samples for Models 2 \& 5).
This limitation remians until cross-task datasets are integrated. As demonstrated by Models 5 \& 6,
% When the optical flow training data reach sufficient scale (\textit{e.g.}, 1M samples to train Model 5), 
adding cross-task data yields minimal improvement for optical flow generalization but significantly boosts cross-task generalization, particularly on the Booster dataset. This improvement stems primarily from the prevalence of non-Lambertian surfaces in Booster, which are underrepresented in existing optical flow simulation datasets. By generating optical flow samples from diverse datasets like Hypersim~\cite{hypersim}, our method effectively generalizes to such specialized scenarios, thereby confirming the value of cross-task data integration.}

\subsubsection{Framework Compatibility to Diverse Encoders}
\label{sec:encoder_ablation}
% 为什么非得是大模型？普通的模型不行吗？顺便证明我们的算法可以应用到CNN-based & Swin-based 的方法中
Our feature transformation pipeline is primarily designed for the plain multi-layer ViT architecture, therefore it is readily applied to recent foundation models, such as DINOv2~\cite{DINOv2}, SAM~\cite{SAM} and DAM~\cite{depth_anything}. 
We validate the compatibility of the adopter with various ViT pretraining weights in Table~\ref{tab:flow_with_different_backbones}, confirming that all variants achieve state-of-the-art generalization performance. 
Notably, we observe that the model using DINOv2-Giant outperform that of using DINOv2-Base.
% , and a similar conclusion can be drawn for models using DAMv2-Large and DAMv2-Base. 
This finding suggests that larger encoder capacity contributes to better zero-shot performance, which aligns with the intuition that larger models possess stronger representation capabilities and more robust priors—both of which are crucial for downstream matching and aggregation processes involved in generalization.

{Moreover, by employing the guided feature upsampling block to flexibly rescale features of arbitrary resolution to the required scale for feature pyramid construction, our transformation module supports diverse encoder architectures with varying output formats, such as multi-scale CNNs and Transformer-based models. 
To validate this compatibility, we employ ConvNeXt-B~\cite{ConvNeXt} and Swin-B~\cite{swin} as frozen encoders, feeding stage-wise outputs into our module.
However, this setup does not outperform the baseline configuration, which fine-tunes the ImageNet-pretrained Twins-SVT-L encoder~\cite{Twins}. 
We attribute this to the use of smaller-scale backbones and the inefficiency of side-tuning compared to direct encoder fine-tuning. 
Nevertheless, direct tuning may be impractical for ViT architectures due to their vast parameter sizes and data demands.}

% \textit{(a) the feature transformation module}, denoted as CNN+FT and Swin+FT, and 
% \textit{(b) one linear projection to each stage and one MLP for multi-stage fusion}, denoted as CNN+Linear and Swin+Linear,
% as the task adapter to obtain pyramid features.
% Although the pyramid architectures like ResNet and Swin encoder preserves the high-resolution features, our feature transformation module outperform the variants using simple projection with more parameters in all metrics, as shown in Table~\ref{}. These results demonstrate that our feature transformation is effective and applicable to diverse architectures.

% 在其他任务领域上, 也有将ViT模型用于稠密任务的, 例如检测和分割, 这些任务的方法用到我们的任务中, 我们的优势在哪, 我们的新颖的地方在哪
% Exploring Plain Vision Transformer Backbones for Object Detection
% On the other hand, recent foundation models, such as DINOv2~\cite{DINOv2}, SAM~\cite{SAM} and DAM~\cite{depth_anything}, adopt plain ViT as the basic network architecture for image encoding. 
% Some efforts on detection and segmentations field also provide thier solutions for utilizing plain ViT for dense prediction. 
% We emperically demonstrate that our matching-specific design is much effective to the ViT adapter on dense matching task and achieve the best performance on the generalization comparison.
% Our feature transformation module is compatiable with various vision foundation models, as shown in Table~\ref{}. These results demostrate the compatibility of feature transformation to different plain ViT pretraineing weights. Moreover, all of these variants achieve the state-of-the-art generalization performance.

We should emphasize that our primary objective in this study is to evaluate the effectiveness and generality of the proposed adopter across different architectures and pretrained weights. 
Given that different pretrained weights stem from varying training strategies and datasets, direct comparisons between pyramid and plain encoders for zero-shot performance would be unfair. Therefore, we omit such comparisons and instead focus on demonstrating the framework's generality. In practice, we adopt DINOv2 as the frozen LVM encoder for PanMatch, as it shows the best generalization performance in our current studies.

\begin{table}[]
    \setlength{\tabcolsep}{3pt}
    \centering
    \caption{Ablation for different encoders. FlowFormer is adopted as the baseline for flow estimation.}
    \begin{tabular}{lllccc}
    \hline
    \multirow{2}{*}{ID} & \multirow{2}{*}{Backbone} & \multirow{2}{*}{Dataset}          & KITTI   & Sintel  & Sintel  \\
                        &                           &                                   & F1-All & clean  & final  \\ \hline
    0                   & Twins-SVT-L~\cite{Twins}  & ImageNet-22k~\cite{imagenet}      & 14.72  & 1.01 & 2.40 \\ \hline
    1                   & ConvNeXt-B~\cite{ConvNeXt}& ImageNet-22k~\cite{imagenet}      & 14.49  & 1.14 & 2.71 \\
    2                   & Swin-B-224$^2$~\cite{swin}& ImageNet-22k~\cite{imagenet}      & 15.89  & 1.20 & 2.65 \\
    3                   & SAM-B~\cite{SAM}          & SAM-2B~\cite{SAM}                 & 14.61  & 1.13 & 2.48 \\
    4                   & DAMv2-B~\cite{DAMv2}      & DA-2B~\cite{DAMv2}                & 12.41  & 0.89 & 2.07 \\
    % 5                   & DAMv2-L~\cite{DAMv2}      & DA-2B~\cite{DAMv2}                & 13.50  & 0.97 & 2.03 \\
    5                   & DINOv2-B~\cite{DINOv2}    & LVM-142M~\cite{DINOv2}            & 13.40  & 1.02 & 2.21 \\
    6                   & DINOv2-G~\cite{DINOv2}    & LVM-142M~\cite{DINOv2}            & 13.39  & 0.82 & 1.94 \\ \hline
    \end{tabular}
    \label{tab:flow_with_different_backbones}
\end{table}

\subsection{Real-World Application}
\label{sec:application}
\begin{figure}[t]
    \centering
    \includegraphics[width=0.5\textwidth]{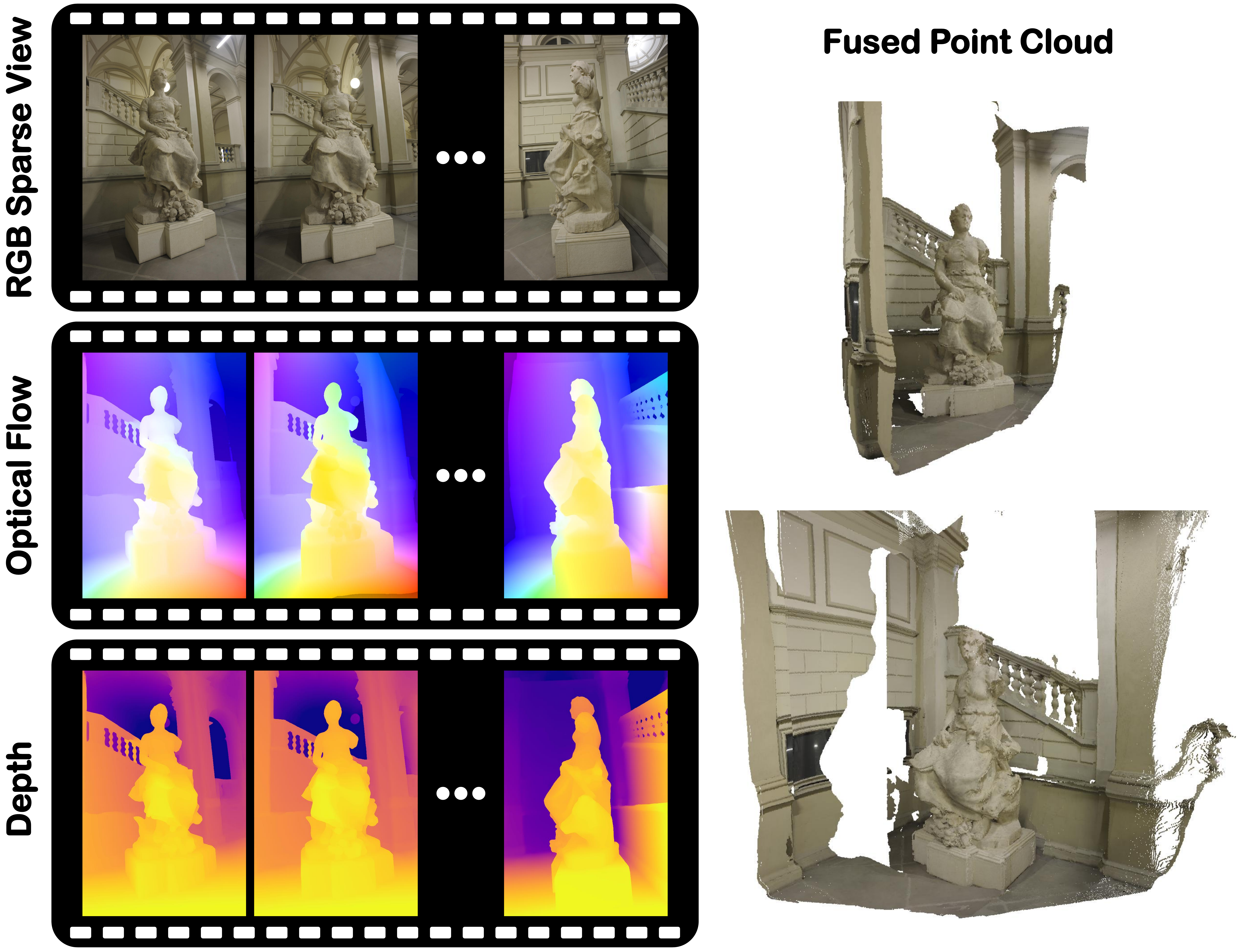}
    \caption{A unique task that the independent methods are unable to solve. Given a traveling video, our model can infer the depth without the need of pose information.
    }
    \label{fig:application}
\end{figure}
Given a video sequence or sparse multi-view images, our method directly estimates per-frame depth maps without relying on pose priors. This is achieved by first generating the unified displacement field. Then, the confidence-filtered dense correspondences are used to estimate relative camera poses, enabling subsequent depth calculation. As illustrated in Fig.~\ref{fig:application}, we back-project these per-frame depth estimates into a common world coordinate system for point cloud registration and fusion. The resulting fused point cloud captures the scene geometry across multiple views, demonstrating geometrically coherent reconstruction. These results validate the robust performance and strong cross-task capabilities of our approach.

\section{Conclusion}
In this work, we present \emph{PanMatch}, a versatile model designed to address diverse correspondence tasks.
This multi-task capability stems from {two key designs: (1) a unified 2D displacement field representation, and (2) a zero-shot generalization strategy implemented by integrating domain-invariant representations from LVMs and pretraining on multi-domain datasets.}
To effectively adapt LVM features for correspondence tasks, we develop a Feature Transformation pipeline that extracts a multi-scale feature pyramid from the LVM representations.
% Additionally, we assemble a multi-domain, multi-task optical flow dataset containing 1.8 million samples to support extensive pretraining.
Furthermore, we reorganize various datasets with task-specific annotation formats (\textit{e.g.}, disparity, depth, and flow) into a common dense displacement field, enabling the aggregation of multi-domain and multi-task datasets for large-scale pretraining.
% We have presented the versatile capacity of PanMatch on multiple 2-frame correspondence tasks including stereo matching, optical flow and feature matching.
% We demonstrated that all 2-frame tasks can be solved within the optical flow formulation. This allows to simplify the challenges from establishing a correspondence matching foundation models to learning domain-invariant features for matching, for which we develop a LVM-equipped feature transformation, in particular regarding the LVM as the lossless domain normalizer to fancilate the cross-domain learning. Our PanMatch is updatable if more datasets and better flow baseline is proposed, revealing its ability to keep up with the times.
Extensive experiments demonstrate that PanMatch achieves state-of-the-art or highly competitive performance across 15 popular benchmarks for flow, stereo, depth and feature matching, underscoring its multitasking versatility and cross-domain generalization.

A key insight of this work is that any two-frame matching task can be effectively addressed under a single paradigm of 2D displacement field prediction. This formulation not only unifies various correspondence matching tasks but also facilitates pretraining with datasets from diverse sources.
We hope our findings will be valuable for advancing dense correspondence and multi-view perception tasks.

\section*{Acknowledge}
This work was partially supported by the National Natural Science Foundation of China (No. 62301601, U20A20185, 62372491), the Guangdong Basic and Applied Basic Research Foundation (2022B1515020103, 2023B1515120087), the Shenzhen Science and Technology Program (No. RCYX20200714114641140).

% use section* for acknowledgment
%\ifCLASSOPTIONcompsoc
%\section*{Acknowledgments}
%\else
% regular IEEE prefers the singular form
%\section*{Acknowledgment}
%\fi

%This work was partially supported by the National Natural Science Foundation of China (No. U20A20185, 61972435).	
	
% \bibliographystyle{unsrt}
\bibliographystyle{IEEEtran}
\bibliography{main}

% Can use something like this to put references on a page
% by themselves when using endfloat and the captionsoff option.
\ifCLASSOPTIONcaptionsoff
\newpage
\fi

% You can push biographies down or up by placing
% a \vfill before or after them. The appropriate
% use of \vfill depends on what kind of text is
% on the last page and whether or not the columns
% are being equalized.

%\vfill

% Can be used to pull up biographies so that the bottom of the last one
% is flush with the other column.
%\enlargethispage{-5in}

% that's all folks
\vspace{-1.5cm}
\begin{IEEEbiography}[{\includegraphics[width=1in,height=1.25in,clip,keepaspectratio]{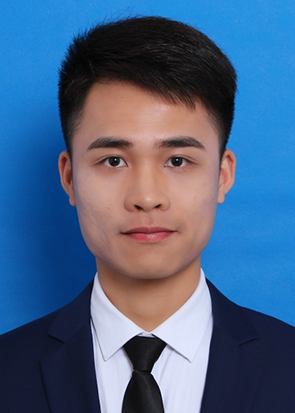}}]
    {Yongjian Zhang} received the B.Eng. degree from
    Sun Yat-sen University (SYSU) in 2021, where he is currently pursuing the Ph.D. degree with the School of Electronics and Communication Engineering. His current research interests focus on stereo matching, depth estimation and completion, and vision-and-language navigation.
\end{IEEEbiography}
\vspace{-1.5cm}
\begin{IEEEbiography}[{\includegraphics[width=1in,height=1.25in,clip,keepaspectratio]{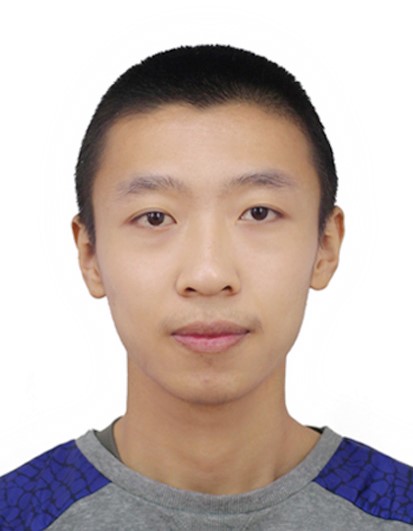}}]
    {Longguang Wang} received the B.E. degree in Electrical Engineering from Shandong University (SDU), Jinan, China, in 2015, and the Ph.D. degree in Information and Communication Engineering from National University of Defense Technology (NUDT), Changsha, China, in 2022. His current research interests include low-level vision and 3D vision.
\end{IEEEbiography}
\vspace{-1.1cm}
\begin{IEEEbiography}[{\includegraphics[width=1in,height=1.25in,clip,keepaspectratio]{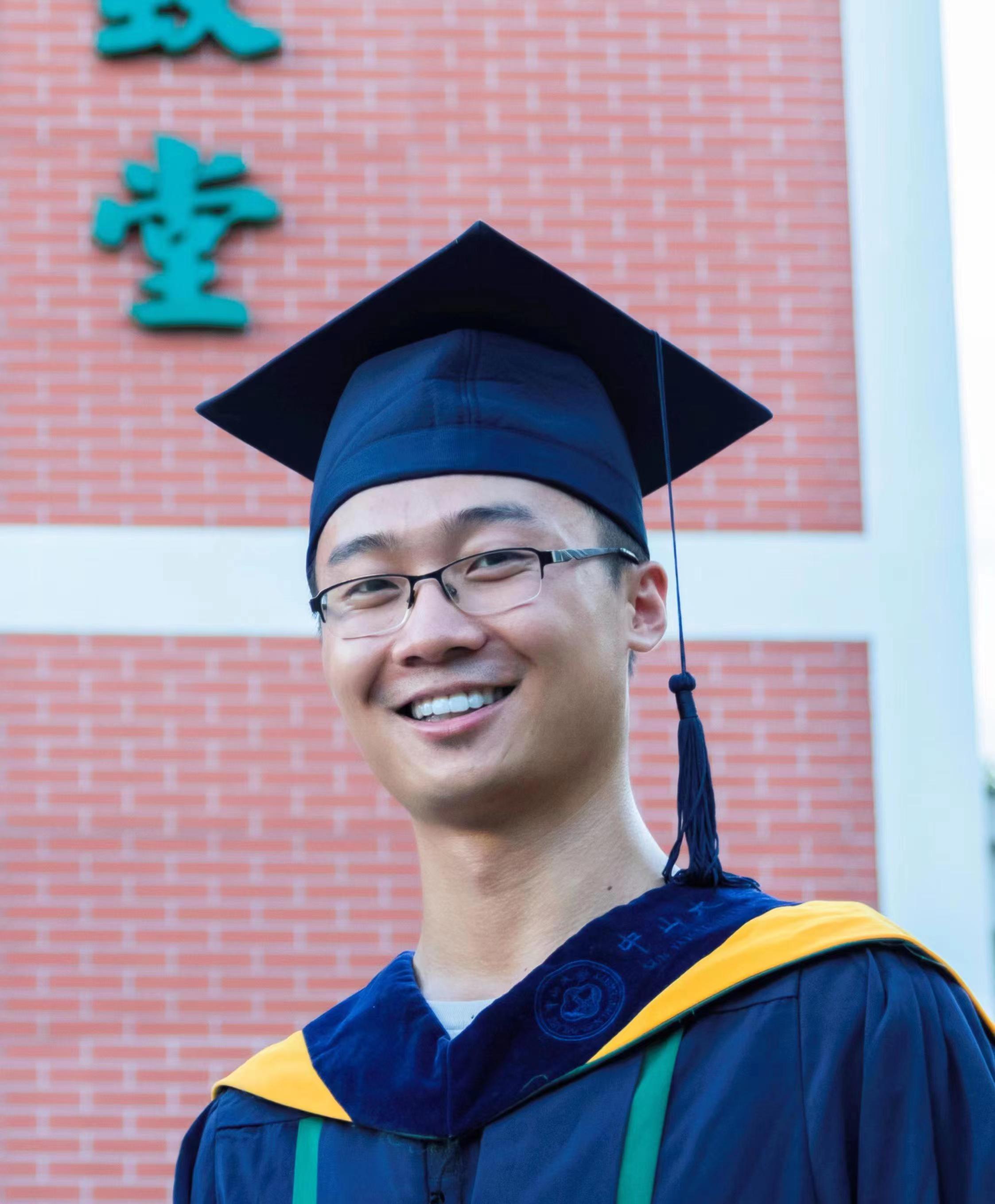}}]
    {Kunhong Li} received the B.Eng. degree from
    Xiamen University (XMU) in 2017 and the M.Eng.
    degree from Sun Yat-sen University (SYSU), China, in 2021, where he is currently pursuing the Ph.D. degree with the School of Electronics and Communication Engineering. His current research interests focus on 3D reconstruction, local feature extraction, and depth estimation.
\end{IEEEbiography}
\vspace{-1.1cm}
\begin{IEEEbiography}[{\includegraphics[width=1in,height=1.25in,clip,keepaspectratio]{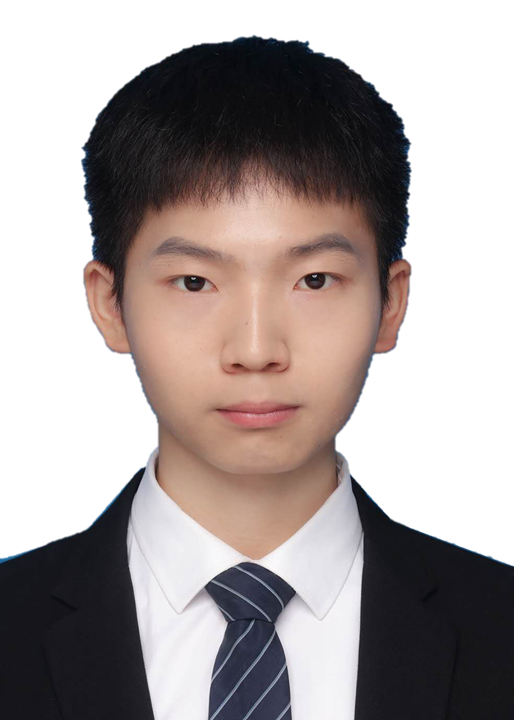}}] 
    {Yun Wang}  received the B.E. degree from China University of Geosciences (CUG) in 2020 and the M.E.
    degree from Sun Yat-sen University (SYSU), China, in 2023.
    He is currently pursuing the Ph.D. degree with the Department of Computer Science, City University of Hong Kong, Hong Kong SAR. His current research interests include deep learning and 3D vision, especially depth estimation. 
\end{IEEEbiography}
\vspace{-1.1cm}
\begin{IEEEbiography}[{\includegraphics[width=1in,height=1.25in,clip,keepaspectratio]{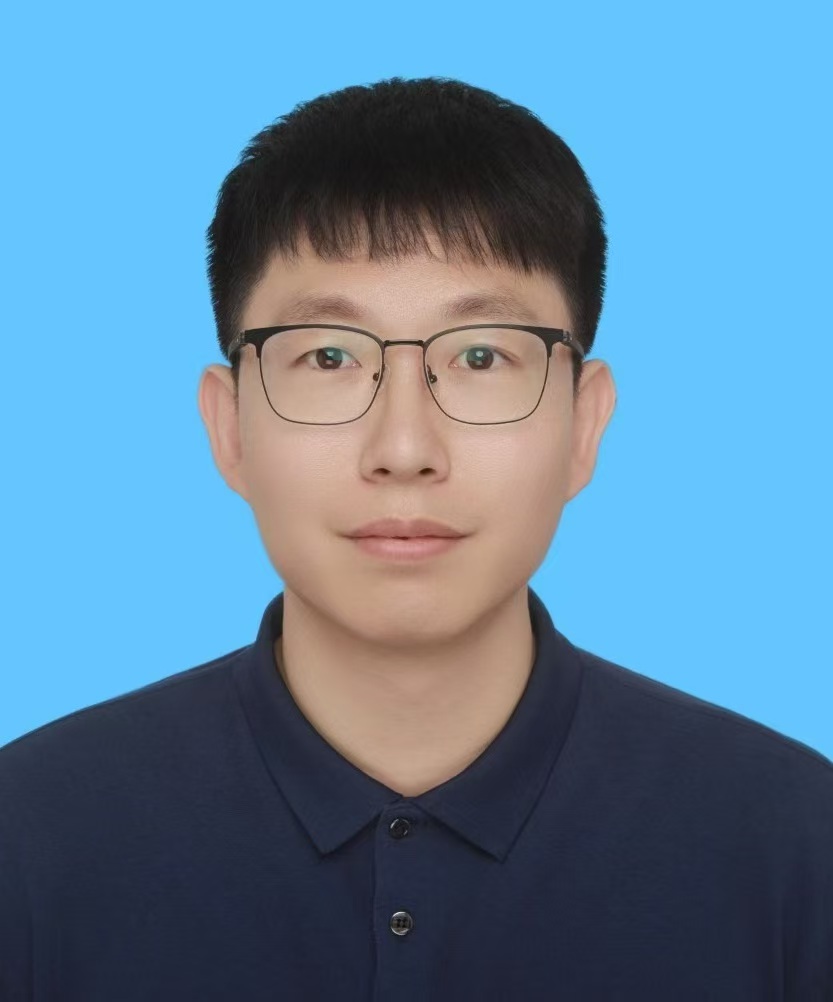}}] 
    {Ye Zhang} received the Ph.D. degree in Electronic Science and Technology from National University of Defense Technology (NUDT), Changsha, China, in 2023. He served as a postdoctor at Sun Yat-sen University (SYSU) from 2023 to 2024. He is currently an associate researcher of the School of Electronics and Communication Engineering, Sun Yat-sen University, Shenzhen, China. His main research interests include 3D vision and human-computer interaction.
\end{IEEEbiography}
\vspace{-1.1cm}
\begin{IEEEbiography}[{\includegraphics[width=1in,height=1.25in,clip,keepaspectratio]{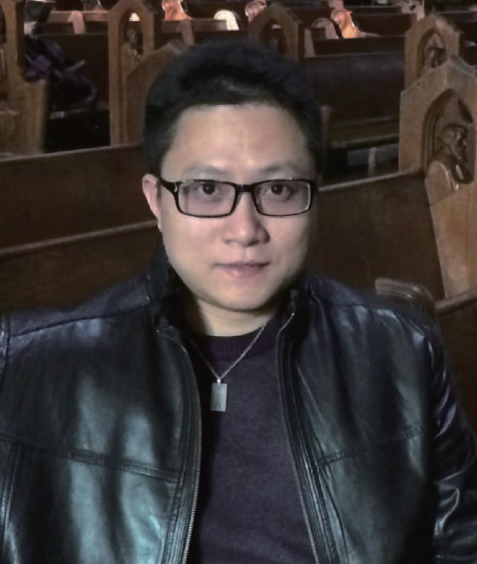}}]
    {Liang Lin} (Fellow, IEEE) is a Full Professor of computer science at Sun Yat-sen University. He served as the Executive Director and Distinguished Scientist of SenseTime Group from 2016 to 2018, leading the R\&D teams for cutting-edge technology transferring. He has authored or co-authored more than 200 papers in leading academic journals and conferences, and his papers have been cited by more than 30,000 times. He is an associate editor of IEEE Trans.Neural Networks and Learning Systems and IEEE Trans. Multimedia, and served as Area Chairs for numerous conferences such as CVPR, ICCV, SIGKDD and AAAI. He is the recipient of numerous awards and honors including Wu Wen-Jun Artificial Intelligence Award, the First Prize of China Society of Image and Graphics, ICCV Best Paper Nomination in 2019, Annual Best Paper Award by Pattern Recognition (Elsevier) in 2018, Best Paper Dimond Award in IEEE ICME 2017, Google Faculty Award in 2012. His supervised PhD students received ACM China Doctoral Dissertation Award, CCF Best Doctoral Dissertation and CAAI Best Doctoral Dissertation. He is a Fellow of IEEE/IAPR/IET.
\end{IEEEbiography}
\vspace{-1.1cm}
\begin{IEEEbiography}[{\includegraphics[width=1in,height=1.25in,clip,keepaspectratio]{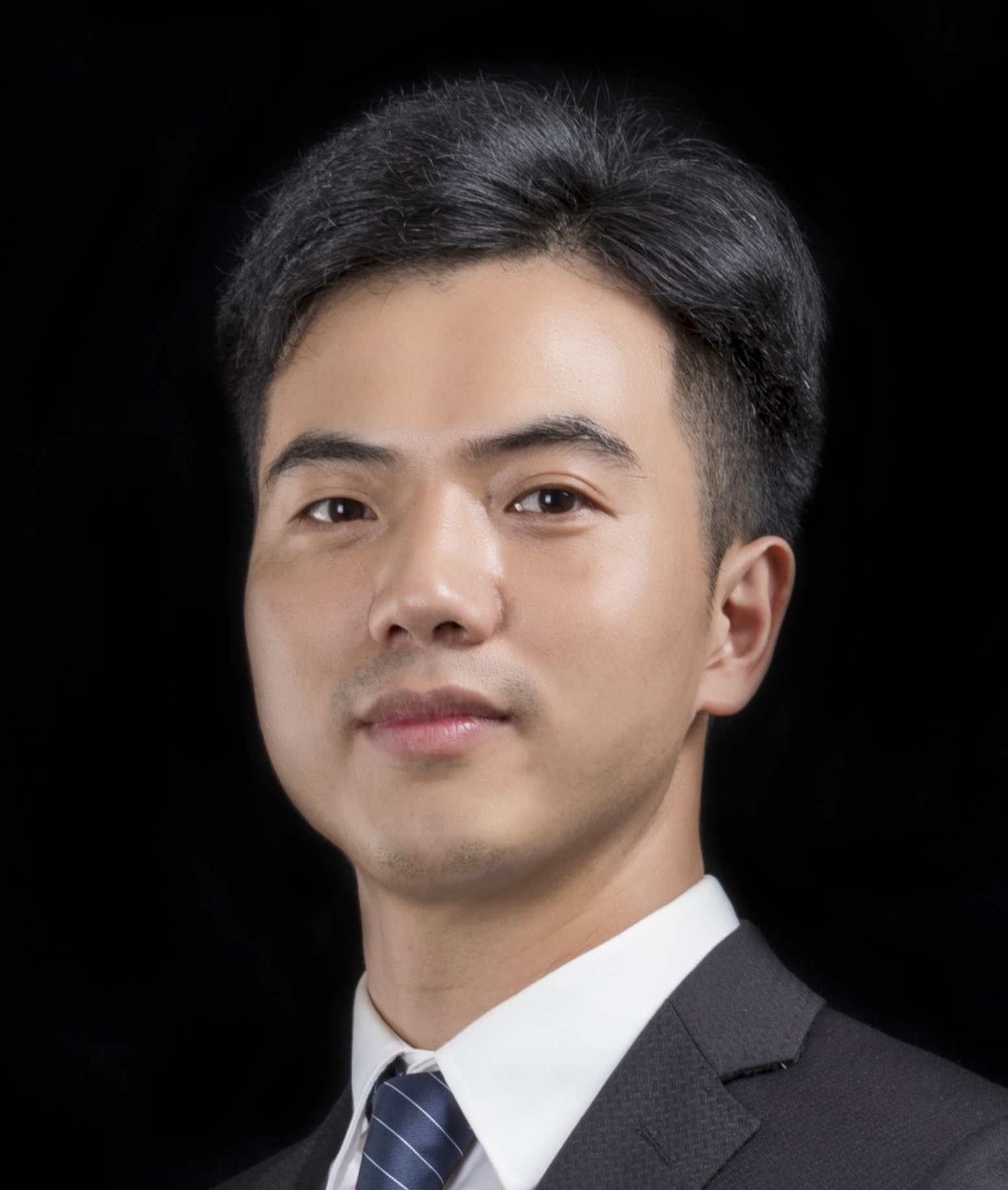}}]
     {Yulan Guo} is a full Professor with Sun Yat-sen University. He has authored over 200 articles at highly referred journals and conferences, receiving over 20,000 citations in Google Scholar. His research interests lie in spatial intelligence, 3D vision, and robotics. He served as a Senior Area Editor for IEEE Transactions on Image Processing, and an Associate Editor for the Visual Computer, and Computers \& Graphics. He also served as an area chair for CVPR 2025/2023/2021, ICCV 2025/2021, ECCV 2024, NeurIPS 2024, and ACM Multimedia 2021. He organized over 10 workshops, challenges, and tutorials in prestigious conferences such as CVPR, ICCV, ECCV, and 3DV. He is a Senior Member of IEEE and ACM.
\end{IEEEbiography}
\end{document}